\documentclass[acmtog,anonymous=False,review=False,nonacm]{acmart}
\renewcommand\footnotetextcopyrightpermission[1]{} 
\acmSubmissionID{1234}

\usepackage{booktabs} 
\usepackage{bbding}
\usepackage{multirow}
\usepackage[ruled]{algorithm2e} 

\SetAlFnt{\small}
\SetAlCapFnt{\small}
\SetAlCapNameFnt{\small}
\SetAlCapHSkip{0pt}

\usepackage{color}
\definecolor{applegreen}{rgb}{0.55, 0.71, 0.0}
\definecolor{autumnorange}{rgb}{0.87, 0.61, 0.33}

\acmJournal{TOG}

\begin{document}
\title{Learning One-Quarter Headshot 3D GANs from a Single-View Portrait Dataset with Diverse Body Poses}

\thanks{Copyright © 2025 IEEE. Personal use of this material is permitted. Permission from IEEE must be obtained for all other uses, in any current or future media, including reprinting/republishing this material for advertising or promotional purposes, creating new collective works, for resale or redistribution to servers or lists, or reuse of any copyrighted component of this work in other works.}

\author{Yiqian Wu}
\affiliation{%
     \institution{State Key Lab of CAD\&CG, Zhejiang University}
     \city{Hangzhou}
     \country{China}
     }
\orcid{0000-0002-2432-809X}
\email{onethousand@zju.edu.cn}

\author{Hao Xu}
\affiliation{%
     \institution{State Key Lab of CAD\&CG, Zhejiang University}
     \city{Hangzhou}
     \country{China}
     }
\orcid{ }
\email{ }

\author{Xiangjun Tang}
\affiliation{%
     \institution{State Key Lab of CAD\&CG, Zhejiang University}
     \city{Hangzhou}
     \country{China}
     }
\orcid{ }
\email{ }
\author{Yue Shangguan}
\affiliation{%
     \institution{University of Texas at Austin}
     \city{AUSTIN}
     \country{United States}
     }
\orcid{ }
\email{ }
\author{Hongbo Fu}
\affiliation{%
     \institution{Hong Kong University of Science and Technology}
     \city{Hong Kong}
     \country{China}
     }
\orcid{ }
\email{hongbofu@cityu.edu.hk}

\author{Xiaogang Jin}
\authornote{Corresponding author.}
\affiliation{%
     \institution{State Key Lab of CAD\&CG, Zhejiang University; ZJU-Tencent Game and Intelligent Graphics Innovation Technology Joint Lab}
     \city{Hangzhou}
     \country{China}
     }
\orcid{0000-0001-7339-2920}
\email{jin@cad.zju.edu.cn}

\makeatletter
\let\@authorsaddresses\@empty
\makeatother

\begin{abstract}

3D-aware face generators are typically trained on 2D real-life face image datasets that primarily consist of near-frontal face data, and as such, they are unable to construct \textit{one-quarter headshot} 3D portraits with complete head, neck, and shoulder geometry. Two reasons account for this issue: First, existing facial recognition methods struggle with extracting facial data captured from large camera angles or back views. Second, it is challenging to learn a distribution of 3D portraits covering the one-quarter headshot region from single-view data due to significant geometric deformation caused by diverse body poses. To this end, we first create the dataset $\it{360}^{\circ}$-\textit{Portrait}-\textit{HQ} (\textit{$\it{360}^{\circ}$PHQ} for short) which consists of high-quality single-view real portraits annotated with a variety of camera parameters (the yaw angles span the entire $360^{\circ}$ range) and body poses. We then propose \textit{3DPortraitGAN}, the first 3D-aware one-quarter headshot portrait generator that learns a canonical 3D avatar distribution from the \textit{$\it{360}^{\circ}$PHQ} dataset with body pose self-learning. Our model can generate view-consistent portrait images from all camera angles with a canonical one-quarter headshot 3D representation. Our experiments show that the proposed framework can accurately predict portrait body poses and generate view-consistent, realistic portrait images with complete geometry from all camera angles. 
We will release our \textit{$\it{360}^{\circ}$PHQ} dataset, code and pre-trained models for reproducible research.

\end{abstract}

\ccsdesc[500]{Computing methodologies~Image-based rendering}

\keywords{Portrait generation, 3D-aware GANs}

\begin{teaserfigure}
  \centering
  \includegraphics[width=0.95\linewidth]{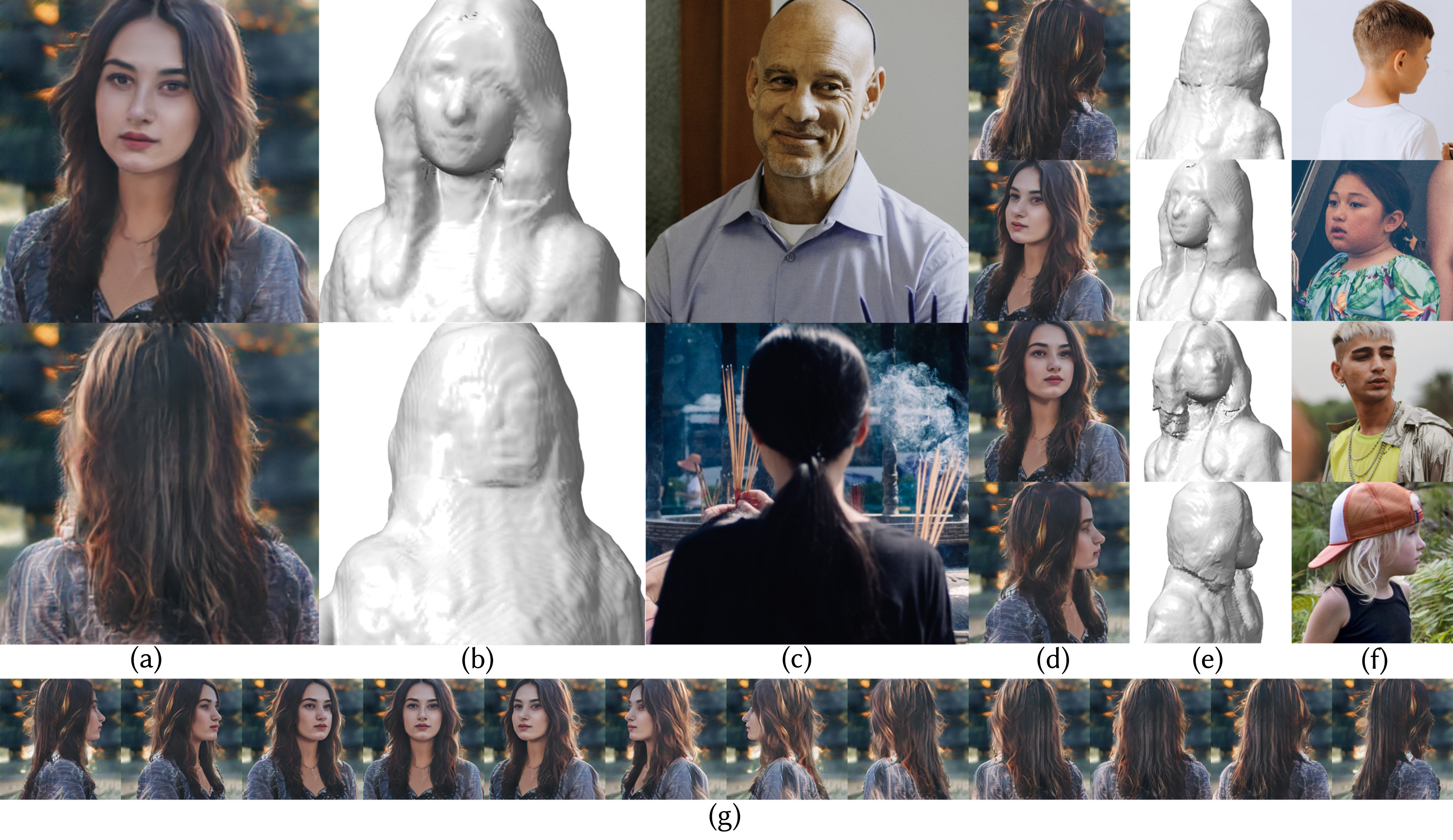}
  \caption{
  Our 3DPortraitGAN can generate one-quarter headshot 3D avatars and output portrait images (a and d) of a single identity using camera poses and body poses from reference images (c and f). The real reference images (c and f) are sampled from our \textit{$\it{360}^{\circ}$PHQ} dataset. Shapes (b and e) are iso-surfaces extracted from the density field of each portrait using marching cubes. We demonstrate that 3DPortraitGAN can generate canonical portrait images from all camera angles by showcasing the ${360}^{\circ}$ yaw angle exploration results in (g).
  }
\label{fig:teaser}
\end{teaserfigure}

\maketitle

\begin{figure*}[t]
  \centering
  \includegraphics[width=\linewidth]{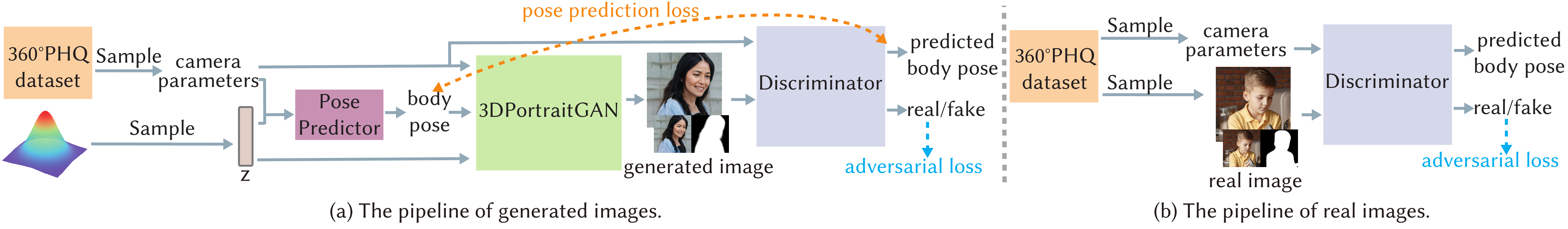}
  
  \caption{
      Our approach's framework and central idea are depicted in this figure, which shows the pipeline for generated images (a) and real images (b).
      To generate images, we propose to use a pose predictor to estimate the body pose distribution under the condition of camera parameters and a latent code. A body pose sampled from the estimated distribution is then utilized to produce a portrait image 
      {(along with its foreground mask)}
      that matches this body pose. We also embed another pose predictor into our discriminator to predict the body pose from the generated or real portrait image.  
      The difference between the predicted and sampled body poses is utilized to train the pose predictor in the discriminator.
      Additionally, the discriminator is conditioned on the predicted body pose to compute scores to train the entire framework.
  }
  \label{fig: motivation}
\end{figure*}
\section{Introduction}
\label{sec: Introduction}

There has been significant progress in the development of 3D-aware generators in recent years.
Unlike 2D GANs, which can only produce high-quality single-view images, 3D-aware generators  \cite{DBLP:conf/cvpr/ChanLCNPMGGTKKW22,DBLP:journals/corr/abs-2112-11427,DBLP:conf/iclr/GuL0T22} utilize voxel rendering or NeRF rendering to acquire knowledge of the 3D geometry from 2D image collections. 3D-aware generators have been instrumental in facilitating image and video editing tasks \cite{DBLP:journals/corr/abs-2205-15517, nerffaceediting,DBLP:conf/siggrapha/JinRKBC22,DBLP:journals/corr/abs-2301-02700,DBLP:journals/corr/abs-2203-13441,DBLP:journals/corr/abs-2302-04871,10.1145/3597300} since they are able to produce multi-view consistent results with realistic geometry.

Most 3D-aware face generators only require single-view face images as training data. These single-view face datasets, such as \textit{FFHQ} \cite{DBLP:conf/cvpr/KarrasLAHLA20} and \textit{CelebA} \cite{liu2015faceattributes}, usually consist of in-the-wild images, which are readily accessible and abundant on the Internet. 
Despite their usefulness, these currently widespread single-view face datasets 
have certain limitations. 
First, the datasets primarily consist of frontal or near-frontal views, with limited views from larger poses and no views from behind the head. 
As a result of using \textit{FFHQ} and \textit{CelebA}, 3D-aware face generators \cite{DBLP:conf/cvpr/ChanLCNPMGGTKKW22,DBLP:journals/corr/abs-2112-11427,DBLP:conf/iclr/GuL0T22,DBLP:conf/nips/SchwarzSNL022} only produce 
frontal-head area data and fail to produce the back of the head. 
Second, although PanoHead \cite{An_2023_CVPR} is able to achieve full-head generation by collecting back-head and large-pose images (\textit{FFHQ-F}) and utilizing self-adaptive image alignment, \textit{FFHQ-F} only includes 
{head}
data and does not contain complete data for the neck and shoulder regions. 
Consequently, the geometry in PanoHead's results is still limited to the facial region, and the neck and shoulders are incomplete.

The reasons behind these dataset limitations are twofold. 
First, the availability of such single-view face datasets mainly depends on the accuracy of face-recognition technology used to extract faces from in-the-wild images and the accuracy of face reconstruction methods utilized to extract camera parameters. In some cases where cameras are positioned at a large camera pose or even behind the head, the necessary facial features required for accurate recognition may be obscured, making it difficult to extract data from all angles.
Second, for in-the-wild \textit{one-quarter headshot}\footnote{\hyperref[]{https://www.backstage.com/magazine/article/types-of-headshots-75557/}} portraits that include the neck and shoulder regions, diverse body poses are always present. Unfortunately, current 3D-aware portrait generators require portrait images in a canonical space where all objects are uniformly positioned, have semantically meaningful correspondence and similar scale, and undergo no significant deformations.
Otherwise, the dimensionality of the data distribution becomes prohibitively high, resulting in significant distortion in the results.
Therefore, the lack of sufficient data and appropriate methods creates a significant challenge for developing a one-quarter headshot 3D-aware portrait generator using a single-view dataset.

Another line of research focuses on multi-view portrait data. 
To reconstruct 3D portrait geometry, researchers have developed 3D portrait datasets \cite{DBLP:conf/cvpr/Yang0WHSYC20,wuu2022multiface,DBLP:journals/corr/abs-1904-00168} that consist of high-quality multi-view labeled portrait images. Nevertheless, the diversity of these datasets is constrained by the challenges involved in data collecting and processing.  
Synthetic portrait datasets \cite{DBLP:journals/corr/abs-2212-06135,DHFdataset,DBLP:conf/iccv/WoodBHD0S21} offer a convenient solution to generate portrait data with diverse environments and camera parameters. 
Given its capacity to regulate data creation and produce reliable ground-truth labels, such as segmentation masks and landmarks, synthetic data has become a popular tool for training computer vision models.
Rodin \cite{DBLP:journals/corr/abs-2212-06135} utilizes rendered portrait images as its training dataset and achieves a one-quarter headshot portrait generation model. However, its results are constrained by the unrealistic rendering style (see Fig. \ref{fig: comparison}) 
of the training data. Additionally, Rodin demands an ample, multi-view image dataset of avatars (consisting of at least 300 multi-view images for each 100K synthetic individuals) to fit tri-planes, making this method highly data-dependent.
In summary, no suitable multi-view face data is available for training a realistic 3D-aware portrait generator.

This paper proposes a novel realistic 3D-aware one-quarter headshot portrait generator, \textbf{3DPortraitGAN}, which can learn a canonical 3D avatar distribution from a collection of single-view real portraits with body pose self-learning. The generator is capable of producing realistic, view-consistent portrait images from $360^{\circ}$ camera angles, including complete head, neck, and shoulder geometry. 
Regarding the training data, we focus on utilizing single-view portrait images from the Internet. Considering the challenges associated with collecting data from all camera angles using face recognition methods, we propose to use more distinctive body features to collect data. 
Specifically, we introduce a new data processing method based on an off-the-shelf body reconstruction method, 3DCrowdNet \cite{DBLP:conf/cvpr/ChoiMPL22}. We apply 3DCrowdNet to extract camera parameters and body poses from in-the-wild images. Then, we identify the desired one-quarter headshot portrait regions to obtain aligned images.
The resulting dataset, named  {$\bf{360^{\circ}}$}-\textbf{P}ortrait-\textbf{HQ} (\textbf{$\bf{360^{\circ}}$PHQ}), comprises \textbf{54,000} high-quality single-view portraits with a wide range of camera angles {(the yaw angles span the entire $360^{\circ}$ range). 
%

{Our framework is built on the backbone of EG3D \cite{DBLP:conf/cvpr/ChanLCNPMGGTKKW22} and incorporates the tri-grid 3D representation and mask guidance from PanoHead \cite{An_2023_CVPR}.}
While we aim to generate human geometry within a canonical space using our generator, the diverse body poses in the \textit{$\it{360}^{\circ}$PHQ} dataset pose a challenge for learning a canonical 3D avatar representation. 
To address this issue, we employ a deformation module to deform the generated human geometry in the canonical space. This ensures that the volume rendering results display the desired body pose to fit within the real portrait distribution.
Since the estimated body poses in the dataset are imprecise, we incorporate two pose predictors into both the generator and discriminator to achieve body pose self-learning.
As depicted in Fig. \ref{fig: motivation}, the generator's pose predictor learns a distribution of body pose, which the generator utilizes to generate portraits. The generated portrait, along with its foreground mask, 
is then processed by the body-pose-aware discriminator, where the pose predictor predicts its body pose. The difference between the estimated and input body poses of the generated portrait is utilized to train the pose predictor in the discriminator.
Additionally, the body-pose-aware discriminator is conditioned on the predicted body pose to score the generated (or real) portraits, which are further used to train the generator and the discriminator networks.

To the best of our knowledge, our delicately designed framework, coupled with the novel portrait dataset, enables our 3DPortraitGAN model to become the first 3D-aware one-quarter headshot GAN capable of effectively learning $360^{\circ}$ canonical 3D portraits from single-view and body-pose-varied 2D data, while also achieving body pose self-learning.
Through extensive experiments, we demonstrate that our framework can generate view-consistent, realistic portrait images with complete geometry from a wide range of camera angles and accurately predict portrait body poses.

In summary, our work makes the following major contributions:
\begin{itemize}

    \item A large-scale dataset of high-quality single-view real portrait images featuring diverse camera parameters and body poses.

     \item The first 3D-aware one-quarter headshot GAN framework that can learn $360^{\circ}$ canonical 3D portrait distribution from the proposed dataset with body pose self-learning. 

\end{itemize}

\section{Related Work}
\label{sec: Related Work}

\subsection{Portrait Image Datasets}
    To enable downstream face applications, such as (conditional) face generation, segmentation, face anti-spoofing, face recognition, and facial manipulation, researchers have collected and processed numerous face image datasets.
    The CelebFaces Attributes Dataset (\textit{CelebA}) \cite{liu2015faceattributes} is a large-scale face attributes dataset with over 200,000 celebrity images and 40 rich attribute annotations. Its variants also provide segmentation masks \cite{CelebAMask-HQ}, spoof \cite{CelebA-Spoof}, and fine-grained labels \cite{jiang2021talkedit} to benefit the community. 
    Since its creation by the authors of StyleGAN \cite{DBLP:conf/cvpr/KarrasLAHLA20}, \textit{FFHQ} has quickly become the most popular dataset for 2D/3D face generation tasks. 
    However, while \textit{CelebA} and \textit{FFHQ} datasets are widely used for training sophisticated 3D-aware generators, their images are predominantly limited to small to medium camera poses. As a result, these generators often produce distorted or incomplete head geometry due to the insufficient range of available camera poses in the training data.
    Although researchers are attempting to incorporate more large-pose data (\textit{LPFF} \cite{DBLP:journals/corr/abs-2303-14407}) and back-head data (\textit{FFHQ-F} \cite{An_2023_CVPR}) into \textit{FFHQ}, these newly added images still lack complete data for the neck and shoulder regions, resulting in incomplete portrait geometry and the inability to generate one-quarter headshot results.

    %
    \textit{300W-LP} \cite{DBLP:conf/cvpr/ZhuLLSL16} is a dataset including 61,225 images across a wide range of camera poses,
    but all the images are artificially synthesized by face profiling.
    \textit{AFLW} \cite{DBLP:conf/iccvw/KostingerWRB11} contains 21,080 face images with large-pose variations, and \textit{LS3D-W} \cite{DBLP:conf/iccv/BulatT17} contains approximately 230,000 images from a combination of different datasets \cite{DBLP:conf/cvpr/SagonasTZP13,DBLP:conf/iccvw/ShenZCKTP15,DBLP:conf/cvpr/ZafeiriouTCDS17,DBLP:conf/cvpr/ZhuLLSL16}. 
    But most images in the above dataset are at low resolution and do not include portraits from the back of the head, limiting their suitability for training high-quality full-head models.

    There exist portrait image datasets that contain multi-view face images.
    Some annotated 3D portrait datasets \cite{DBLP:conf/cvpr/Yang0WHSYC20,wuu2022multiface,DBLP:journals/corr/abs-1904-00168} offer high-quality multi-view portrait images and accurate camera parameters suitable for 3D face reconstruction. However, the limited variety in these datasets poses challenges for researchers seeking more diverse data.
    Synthetic portrait datasets \cite{DBLP:journals/corr/abs-2212-06135,DHFdataset,DBLP:conf/iccv/WoodBHD0S21} offer a convenient solution for computer vision tasks, as users can easily control data generation and obtain ground truth labels by using graphics. Despite this advantage, synthetic portrait datasets still have a big domain gap with real-world data, making it challenging to apply these datasets in practical applications.
    
    In summary, no dataset adequately provides high-quality, diverse, and realistic portrait images with a full range of camera poses to train a one-quarter headshot 3D-aware generator. 
    In this paper, we propose a large-scale dataset of high-quality single-view real portrait images featuring diverse camera parameters and body poses and covering the one-quarter headshot region. This allows us to train a one-quarter headshot 3D-aware generator.

\subsection{3D-aware Generators}
    Since Goodfellow et al.'s seminal proposal of generative adversarial networks (GANs) \cite{DBLP:conf/nips/GoodfellowPMXWOCB14} in 2014, numerous GAN models \cite{DBLP:journals/corr/RadfordMC15,DBLP:conf/nips/GulrajaniAADC17,DBLP:conf/iclr/BrockDS19,DBLP:conf/iclr/KarrasALL18} have been developed to achieve remarkable performance in realistic image synthesis.
    The scope of 2D generators has been extended to encompass 3D multi-view rendering these years.   
    Early techniques combine voxel rendering \cite{DBLP:conf/nips/Nguyen-PhuocRMY20,DBLP:conf/nips/ZhuZZ00TF18} or NeRF rendering \cite{DBLP:conf/iclr/GuL0T22,DBLP:conf/cvpr/ChanMK0W21} with generators to facilitate view-consistent image synthesis.
    
    The 3D representations inside 3D-aware GANs could be parameterized by coordinate-based networks \cite{DBLP:conf/cvpr/ChanMK0W21,DBLP:conf/iclr/GuL0T22}, feature maps \cite{DBLP:conf/cvpr/ChanLCNPMGGTKKW22}, signed distance fields \cite{DBLP:conf/nips/0004SWCYLLGF22}, or voxel grids \cite{DBLP:conf/nips/SchwarzSNL022,DBLP:conf/nips/ZhuZZ00TF18}.
    To reduce the cost of calculation, several studies propose the use of a super-resolution network to enhance image quality \cite{DBLP:conf/cvpr/ChanLCNPMGGTKKW22,DBLP:conf/siggraph/TanFMOTPTTZ22,DBLP:journals/corr/abs-2112-11427,DBLP:conf/cvpr/XueLSL22}. 
    Moreover, researchers suggest an efficient optimization strategy \cite{DBLP:journals/corr/abs-2206-10535} to directly generate high-resolution results without using a super-resolution module.
    To eliminate the dependency on 3D pose priors, researchers also propose a pose-free training strategy \cite{shi2023pof3d} to learn the pose distribution of the dataset by the network itself.

    To apply 3D-aware generators to real image and video editing, novel inversion methods \cite{DBLP:journals/corr/abs-2203-13441,DBLP:conf/wacv/KoCCRK23,xie2022high,DBLP:journals/corr/abs-2211-16927,DBLP:journals/corr/abs-2302-04871} have been proposed to obtain latent codes of given input images. 
    Thanks to the capability of 3D-aware GANs to generate view-consistent results and realistic facial geometry, downstream applications such as semantic editing \cite{DBLP:journals/corr/abs-2205-15517,nerffaceediting,DBLP:journals/corr/abs-2302-04871} and portrait stylization \cite{DBLP:conf/siggrapha/JinRKBC22,DBLP:journals/corr/abs-2301-02700} have achieved remarkable performance.

    {
    However, most methods within the face domain utilize datasets containing frontal or near-frontal views, such as \textit{FFHQ} and \textit{CelebA}, leading to incomplete head geometry.
    PanoHead \cite{An_2023_CVPR} trained on \textit{FFHQ-F}, an augmented variant of \textit{FFHQ}, is able to generate full-head results. However, since \textit{FFHQ-F} only covers the head area, PanoHead can only generate the head area, and its results lack the complete geometry of the neck and shoulders. 
    Although some methods for tackling the aforementioned problems are available, such as Rodin \cite{DBLP:journals/corr/abs-2212-06135} and HeadNeRF \cite{DBLP:journals/corr/abs-2112-05637}, which can learn 3D avatar head representations and even generate neck and shoulder data, they rely on multi-view portrait datasets in a canonical space, making them highly dependent on data.

    Ultimately, no appropriate method for learning portrait geometry from in-the-wild single-view portrait images is available, particularly when the data is in a deformed space due to diverse body poses.
    To address this problem, we propose a novel framework to learn the $360^{\circ}$ canonical 3D portrait distribution from a body-pose-varied dataset.

\subsection{Deformable Neural Radiance Fields}
    The continuous, volumetric representation for rendering objects and scenes proposed by Neural Radiance Fields (NeRF) \cite{DBLP:conf/eccv/MildenhallSTBRN20} has benefited the entire graphics and vision community. 
    However, NeRF is limited to static scene rendering.
    To address this limitation, a line of approaches has been explored to achieve non-rigid scenes by deforming sample points in the observation space to the canonical space before querying a template NeRF. 
    Neural networks, such as MLP, are used to represent the continuous deformation of each sample point by outputting the deformation field value.
    %
    In addition to using point coordinates as input, the NN-based deformation field could use time steps \cite{DBLP:conf/cvpr/PumarolaCPM21,DBLP:journals/corr/abs-2303-14435}, latent codes \cite{DBLP:conf/iccv/ParkSBBGSM21,DBLP:conf/iccv/TretschkTGZLT21}, or view and facial expressions \cite{DBLP:conf/cvpr/GafniTZN21}.
    HumanNeRF \cite{DBLP:conf/cvpr/WengCSBK22} further divides the deformation field into skeletal rigid and non-rigid motion to enhance human animation rendering.
    Some works use blend skinning \cite{DBLP:conf/cvpr/YangVNRVJ22,DBLP:conf/cvpr/ZhengABCBH22} or deformable meshes \cite{DBLP:conf/cvpr/AtharXSSS22,DBLP:journals/corr/abs-2304-05097} as guidance instead of encoding the entire deformation field into a neural network.
    To accommodate the representation of discontinuous topological changes, HyperNeRF \cite{DBLP:journals/tog/ParkSHBBGMS21} raises NeRFs into a hyper-space to represent the radiance field corresponding to each input frame as a slice through this space.
    Generative NeRFs also make use of deformation fields based on deformable meshes like 3DMM \cite{DBLP:conf/siggraph/BlanzV99, DBLP:conf/avss/PaysanKARV09}, SMPL \cite{DBLP:journals/tog/LoperM0PB15} and FLAME \cite{DBLP:journals/tog/LiBBL017} to achieve animated results \cite{DBLP:journals/corr/abs-2211-11208,DBLP:conf/nips/BergmanKWCLW22,DBLP:conf/nips/SunWHZW022}.
    {
    For our purposes, we adopt and refine the mesh-guided deformation from RigNeRF \cite{DBLP:conf/cvpr/AtharXSSS22}, which is readily available and has adjustable parameters.
    }

\begin{figure}[t]
  \centering
  \includegraphics[width=\linewidth]{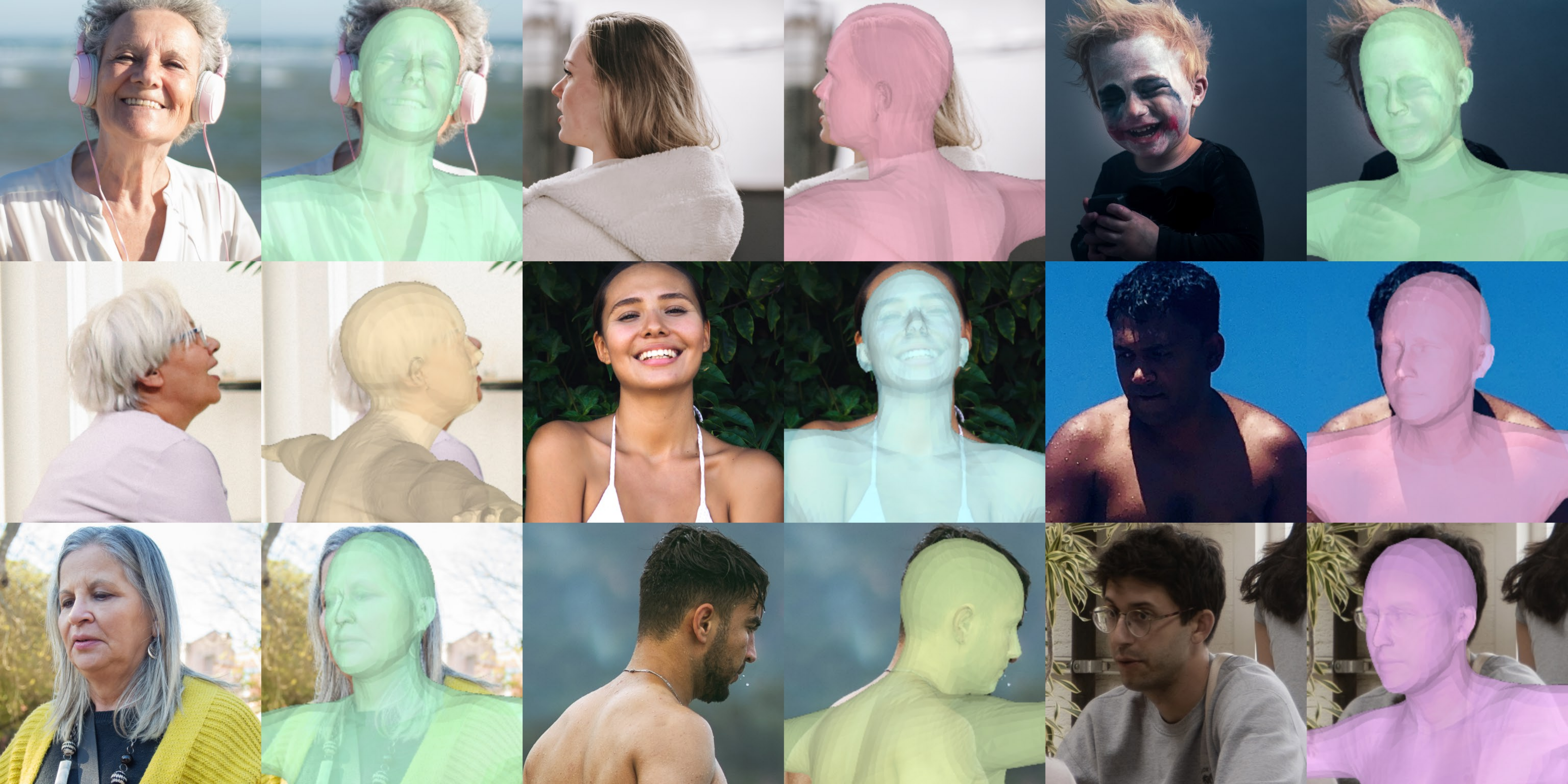}
  
  \caption{
  Random samples from our \textit{$\it{360}^{\circ}$PHQ} dataset. 
  {We show the extracted camera parameters $c$ and body pose $p$ for each image by rendering the SMPL mesh $M(p)$ from $c$.}
  }
  \label{fig: dataset_sample}
\end{figure}

\section{Dataset}
\label{sec: Dataset}
In this section, we will describe our data processing pipeline. We focus on utilizing single-view portrait images online due to their easy accessibility and abundance. However, using face recognition and face reconstruction methods to gather annotated data from all camera angles is challenging since the facial features required for accurate recognition may be obscured. Therefore, we propose {using more distinctive} body features (e.g., shoulders) to collect data.
In particular, we introduce a novel data processing method based on an off-the-shelf body reconstruction method \cite{DBLP:conf/cvpr/ChoiMPL22} to extract camera parameters and body poses from in-the-wild images, enabling us to obtain aligned portraits.

We begin by assuming that we have a human body SMPL \cite{DBLP:journals/tog/LoperM0PB15} template mesh in the local space, denoted as $M$, with the standard body shape. 
Its neck joint is aligned to the origin point $[0,0,0]$, and no additional global rotation or translation is performed on $M$. We denote the template mesh with body pose parameters 
$\vec{\theta} \in \mathbb{R}^{69}$ as $M(\vec{\theta})$.
As we aim to solely preserve the head, neck, and shoulder regions of the input portrait, we only consider the neck pose $p_n \in \mathbb{R}^{3}$ and head pose $p_h \in \mathbb{R}^{3}$ in $\vec{\theta}$, 
{while setting the remaining body pose as zero.}
Thus, we define the neck pose and head pose as $p = [p_n,p_h] \in \mathbb{R}^{6}$ and denote the template mesh with neck and head pose $p$ as $M(p)$.
Regarding camera settings, similar to EG3D \cite{DBLP:conf/cvpr/ChanLCNPMGGTKKW22}, we assume that our camera is always positioned on a sphere with radius $r = 2.7$, directed towards a fixed point. Additionally, intrinsic camera parameters are fixed as constant values. 

Given an in-the-wild portrait image, our aim is to find the camera parameters, neck pose, and head pose of the portrait, 
{allowing us to render the mesh $M$ aligning with the input portrait's head, neck, and shoulder regions.}
Using an off-the-shelf body reconstruction method, 3DCrowdNet \cite{DBLP:conf/cvpr/ChoiMPL22}, we extract the input portrait's SMPL parameters (global rotation $rot$ and translation $trans$, shape parameters $\vec{\beta} \in \mathbb{R}^{10}$, and pose parameters $\vec{\theta} \in \mathbb{R}^{69}$), resulting in an estimated mesh $\tilde{M}(trans, rot, \vec{\beta},\vec{\theta})$ in the world space (with a fixed camera).
%
{We set the neck pose and head pose of $M$ to be the same as those of the estimated mesh}, 
resulting in $M(p)$. Then we compute the transformation matrix that could transform $\tilde{M}(trans, rot, \vec{\beta},\vec{\theta})$ to align its head, neck, and shoulders joints with those of $M(p)$. 
Next, we apply the same transformation matrix to the fixed camera and normalize its camera parameters according to our camera assumption, obtaining the final camera parameters. 
The final camera parameters, denoted as $c \in \mathbb{R}^{25}$, comprise an extrinsic camera matrix $e \in \mathbb{R}^{16}$ and an intrinsic camera matrix $k \in \mathbb{R}^{9}$. Note that $k$ remains fixed as a constant matrix.
Then the raw image is cropped and aligned based on the obtained parameters. To ensure that the head, neck, and shoulders are fully covered, we set a uniform crop region for all images. This process results in an aligned one-quarter headshot image denoted as $I$ (see Fig. \ref{fig: dataset_sample}). 
%

\begin{figure}[t]
  \centering
 \includegraphics[width=0.8\linewidth]{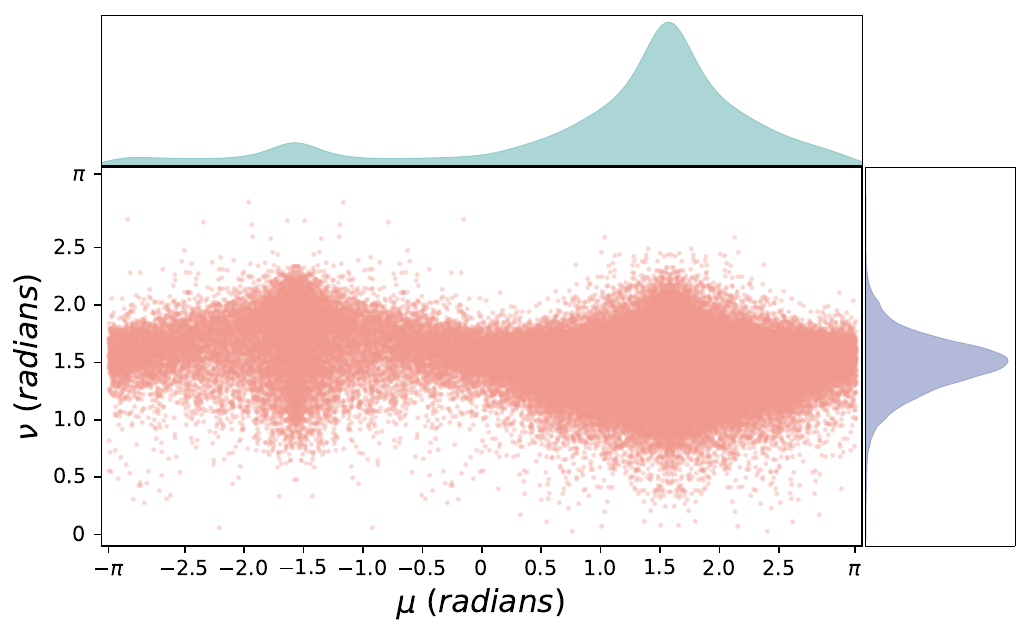}
  
  \caption{The distribution of camera positions of our \textit{$\it{360}^{\circ}$PHQ} dataset.
  Notice that $\mu > 0$ refers to the front view, while $\mu < 0$ refers to the rear view.
  }
  \vspace{-5pt}
  \label{fig: data_distribution}
\end{figure}

To filter out the images with inaccurately estimated camera parameters, we render the mesh $M(p)$ on $I$ using the camera parameters $c$. We then manually examine the rendering results and remove any images where the mesh rendering is not well-aligned with $I$, as well as blurry or noisy images.
However, due to the limitations of the body reconstruction method, we encounter cases where the neck and head rendering results are not aligned with $I$, even though shoulder reconstruction is more accurate. During manual selection, we avoid using the neck and head alignment as a criterion for manual selection and only consider the shoulder alignment.
As a result, the estimated camera parameters can render the template mesh in the local space to have aligned shoulders with the portraits. However, the estimated neck and head pose $p$ is ``coarse'' and inaccurate.
Therefore, instead of being used directly as the training label, this coarse body pose is only employed to calculate a regularization loss during the early stages of the training process, which we will explain in later sections (Sec. \ref{sec: Body Pose Regularization Loss}).

In sum, we collected 41,767 raw portrait images from Pexels\footnote{\hyperref[]{https://www.pexels.com}} and Unsplash\footnote{\hyperref[]{https://unsplash.com}} and finally got \textbf{54,000} aligned images as our \textit{$\it{360}^{\circ}$PHQ} dataset. The number of aligned images is greater than that of raw images, since a single raw image may contain multiple people.  
Samples of these images can be found in Fig. \ref{fig: dataset_sample} as well as the supplementary file {(Sec. E)}.
Our dataset is augmented by a horizontal flip.
We also extract the portrait {foreground segmentation mask} 
from each aligned image using the DeepLabV3 ResNet101 network \cite{DBLP:journals/corr/ChenPSA17}. These segmentation masks will be used as training guidance, as explained in Sec. \ref{sec: Discriminator}.
The images in our dataset are of high quality, with variations in gender, age, race, expression, and lighting. The analysis of the distribution of semantic attributes (gender, race, age, etc) can be found in the supplementary file (Sec. D). 

We convert the camera positions in our dataset to the spherical coordinate system ($\mu$ and $\nu$, or yaw and pitch), and visualize the distribution of camera positions in Fig. \ref{fig: data_distribution}. 
Our dataset contains a diverse set of camera distributions, {with yaw angles spanning the entire $360^{\circ}$ range.}

\section{Methodology}
\label{sec: Methodology}
We first present a comprehensive overview of 
{this section}. 
Firstly, we introduce the design of our body pose-aware discriminator in Sec. \ref{sec: Discriminator}, which is capable of extracting body poses and corresponding scores from the input images.
Subsequently, we elaborate on our 3DPortraitGAN in  Sec. \ref{sec: 3DPortraitGAN}, including the pose predictor in generator (Sec. \ref{sec: pose sampling}), the generator backbone (Sec. \ref{sec: EG3D Backbone}) and our deformation module (Sec. \ref{sec: Deformation Module}).
Finally, we discuss the losses utilized in our training process in  Sec. \ref{sec: Losses}, along with the details of the training in  Sec. \ref{sec: Training Details}.

\subsection{Body Pose-aware Discriminator}
\label{sec: Discriminator}
In Sec. \ref{sec: Dataset}, we mentioned that the body poses in our dataset are inaccurate and cannot be directly used for training 
{(see Sec. \ref{sec: ablation pose prediction})}. 
Inspired by the pose-free 3D-aware generator, Pof3D \cite{shi2023pof3d}, we propose employing the discriminator to predict more accurate body pose $\hat{p}$ from real/generated images.

Taking a real image $I_{real}$ as an example, we denote its camera parameters in the \textit{$\it{360}^{\circ}$PHQ} dataset as $c_{real}$. 
{In EG3D \cite{DBLP:conf/cvpr/ChanLCNPMGGTKKW22}, the dual discrimination requires feeding low- and high-resolution images into the discriminator, resulting in $I_{real}$ consisting of two parts: low-resolution $I_{real}^{-}$ and high-resolution $I_{real}^{+}$ (the former is bilinearly upsampled to the same resolution as the latter). 
Similar to PanoHead \cite{An_2023_CVPR}, we further include the foreground mask $I_{real}^{seg}$ as an input to our discriminator, resulting in:
\begin{equation}
    \begin{split}
    \label{eqn: I_real}
    I_{real} = [I_{real}^{-}, I_{real}^{+}, I_{real}^{seg}],
    \end{split}
\end{equation}
where $I_{real}^{+}$, $I_{real}^{-}$, and $I_{real}^{seg}$ are concatenated into a seven-channel image $I_{real}$. We obtain $I_{real}^{seg}$ using an off-the-shelf network as mentioned in Sec. \ref{sec: Dataset}.
We find that this mask-aware design not only disentangles the foreground from the background, but also helps eliminate the ``rear-view-face'' artifact, which is the presence of a face on the back of the head (as seen in Fig. \ref{fig: rear-view-reg} in ablation studies). This is due to the 3D prior provided by the foreground segmentation masks to some extent.}

As shown in Fig. \ref{fig: discriminator}, the convolutional layers first extract features from $I_{real}$. Then the features and the camera parameters are fed into a pose predictor branch $\Gamma_D$, yielding a predicted body pose:
\begin{equation}
	\begin{split}
    \label{eqn: pose_predict_D}
     \hat{p}_{real} &= \Gamma_D (\mathrm{Conv}(I_{real}), c_{real}),
     \end{split}
\end{equation}
where $\mathrm{Conv}$ denotes the convolutional layers.
We observe that $\Gamma_D$ faces difficulty accurately predicting symmetry poses for symmetry images. To explicitly maintain the symmetry of $\Gamma_D$, we propose an explicit symmetry strategy for $\Gamma_D$.
Specifically, once fed into $\Gamma_D$, we flip the input image horizontally when the spherical coordinate $\mu$ of its camera position falls within the range of $[-\frac{1}{2}\pi,\frac{1}{2}\pi]$ (which indicates that the camera is on the right-hand side of the subject, see Fig. \ref{fig: data_distribution}). We also flip the resulting predicted body pose so as to obtain the final predicted body pose. This operation guarantees that two {horizontally} symmetrical images have the symmetry value of the predicted body pose.

\begin{figure}[t]
  \centering
  \includegraphics[width=\linewidth]{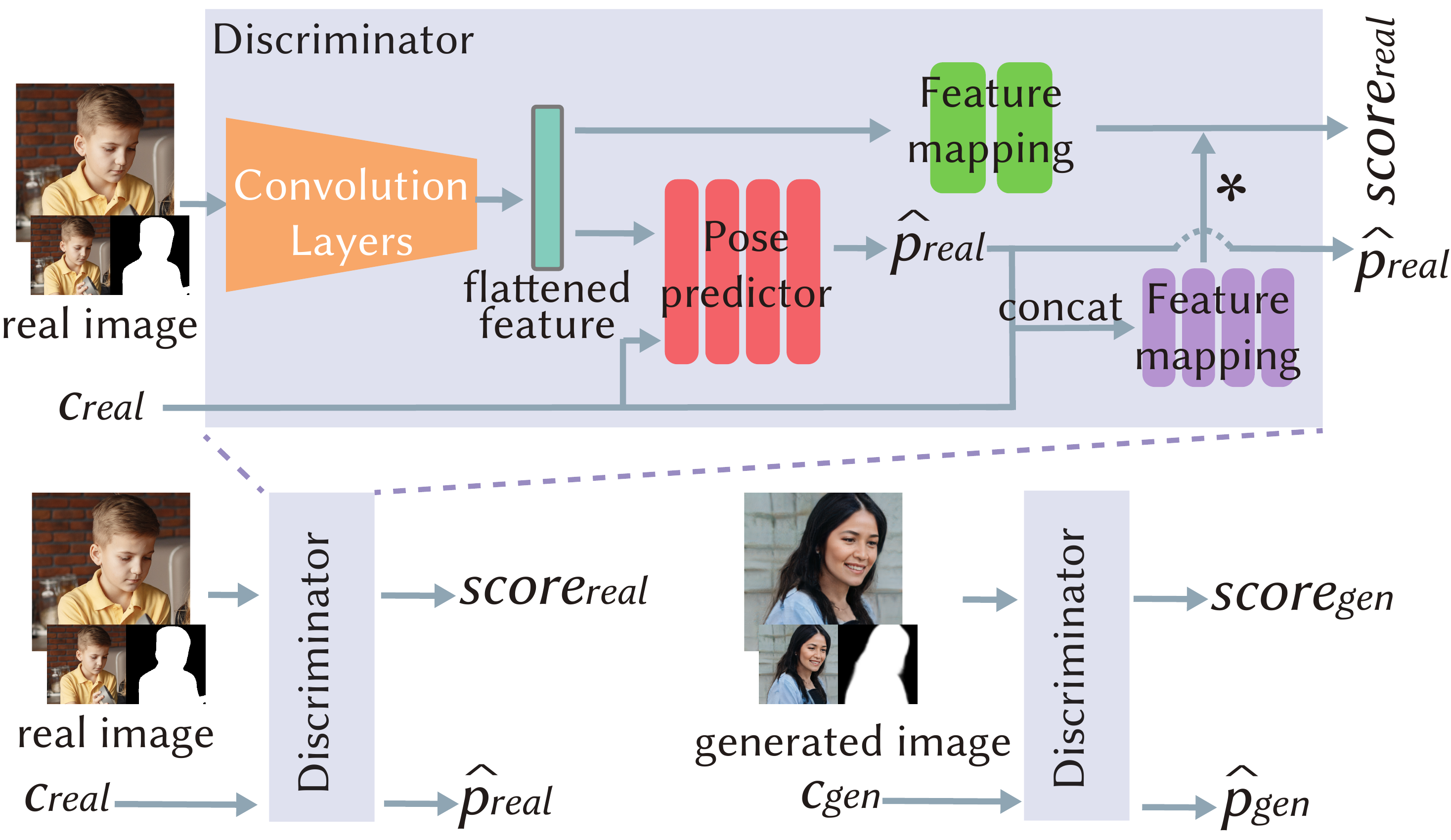}
  
  \caption{The architecture of our body pose-aware discriminator.
    The convolutional layers extract features from the input image {(either a real or generated image)}, which are subsequently fed into a pose predictor along with the camera parameters to predict the body pose.
    The image features, camera parameters, and predicted body pose are subsequently fed into feature mapping networks to acquire the image score.
  }
  \vspace{-15pt}
  \label{fig: discriminator}
\end{figure}

Next, we feed $c_{real}$ and $\hat{p}_{real}$ into a pose feature mapping network, and the image features are fed into an image feature mapping network. Then the outputs of the two feature mapping networks are multiplied to get the final score of the discriminator: 
\begin{equation}
	\begin{split}
    \label{eqn: discriminator_real}
     score_{real} & = \Phi_{image} (\mathrm{Conv}(I_{real})) \cdot \Phi_{pose} (c_{real}, \hat{p}_{real})   \\
     & = D(I_{real}\vert c_{real}, \hat{p}_{real} ), 
     \end{split}
\end{equation}
where $D$ denotes the discriminator, and $\Phi_{image}$ and $\Phi_{pose}$ denote the image feature mapping network and the pose feature mapping network, respectively.

Likewise, $D$ estimates scores and body poses from generated images as:
\begin{equation}
	\begin{split}
    \label{eqn: discriminator_gen}
    I_{gen}  &= [I_{gen}^{-}, I_{gen}^{+}, I_{gen}^{seg}], \\
    \hat{p}_{gen} &= \Gamma_D (\mathrm{Conv}(I_{gen}), c_{gen}), \\
     score_{gen} & = D(I_{gen}\vert c_{gen}, \hat{p}_{gen} ),
     \end{split}
\end{equation}
%
{where $I_{gen}^{-}$ and $I_{gen}^{+}$ respectively refer to the low- and high-resolution images generated by our generator. Additionally, $I_{gen}^{seg}$ is rendered using volume rendering, as detailed in Sec. \ref{sec: EG3D Backbone}.}
$c_{gen}$ is sampled from the \textit{$\it{360}^{\circ}$PHQ} dataset, while $\hat{p}_{gen}$ denotes the predicted body pose of $I_{gen}$.

\begin{figure*}[t]
  \centering
  \includegraphics[width=\linewidth]{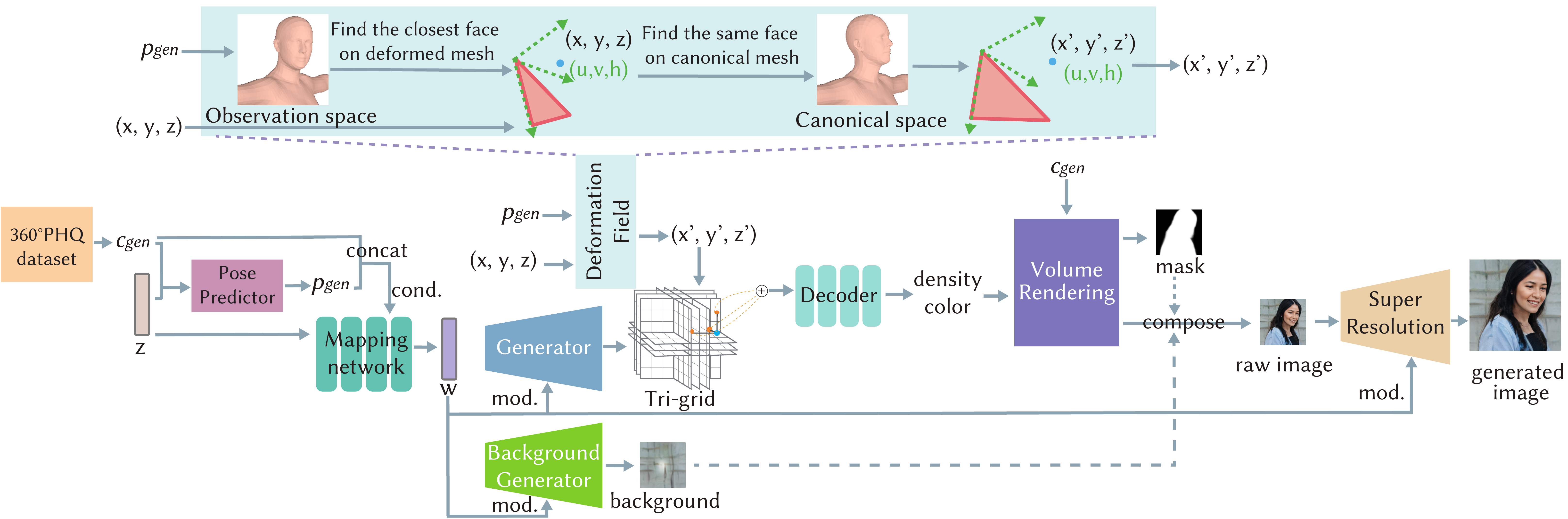}
  
  \caption{
    {Our image generation pipeline}.
    The $z$ latent code sampled from the Gaussian distribution and the camera parameters $c_{gen}$ sampled from the \textit{$\it{360}^{\circ}$PHQ} dataset are fed into a pose predictor to acquire the body pose $p_{gen}$.
    Then 
    $p_{gen}$ and 
    $c_{gen}$ serve as the conditional labels of the mapping network to map the $z$ latent code to the $w$ latent code. Consequently, the $w$ latent code modulates the main generator to generate a tri-grid {and modulates the background generator to generate a background image}.
    During volume rendering, we use the body pose $p_{gen}$ to produce a deformed mesh, which can be utilized to compute a deformation field to generate a portrait image { and a foreground mask that matches $p_{gen}$.
    Then the rendered 
    and 
    background images are composed using the foreground mask. The composed raw image is} 
    subsequently fed into the super-resolution module to acquire the final result.
  }
  \label{fig: pipeline}
\end{figure*}

\subsection{3DPortraitGAN}
\label{sec: 3DPortraitGAN}

\subsubsection{Pose Sampling}
\label{sec: pose sampling}
In this paper, we propose to generate $I_{gen}$ using a latent code $z$, camera parameters $c_{gen}$ and a certain body pose $p_{gen}$ as :
\begin{equation}
	\begin{split}
    \label{eqn: generate}
     I_{gen} = G(z, c_{gen}, p_{gen}) \sim P(I_{gen} \vert z, c_{gen}, p_{gen}).
     \end{split}
\end{equation}
To train our generator, we need to sample camera parameters and body poses from the pose distribution of the \textit{$\it{360}^{\circ}$PHQ} dataset.

However, in the \textit{$\it{360}^{\circ}$PHQ} dataset, while the camera parameters we extract from real images are somewhat precise (after our manual selection), the body poses are ``coarse'' and cannot be used as training data (Sec. \ref{sec: Dataset}).
Taking inspiration from Pof3D \cite{shi2023pof3d}, we employ a pose prediction network $\Gamma_G$ to estimate the conditional distribution of body poses based on randomly sampled latent codes and camera parameters drawn from the \textit{$\it{360}^{\circ}$PHQ} dataset. More specifically, we employ a predictor $\Gamma_G$  to predict a body pose from the 
latent code and camera parameters as follow:
\begin{equation}
	\begin{split}
    \label{eqn: pose_predict_G}
     {p}_{gen} &= \Gamma_G (z, c_{gen})  \sim P({p}_{gen} \vert z, c_{gen}).
     \end{split}
\end{equation}
Similarly to $\Gamma_D$, we want the predicted body poses that correspond to symmetry camera poses to have symmetry distribution. Thus we apply the explicit symmetry strategy to $\Gamma_G$.
We horizontally flip the camera parameters $c_{gen}$ with $\mu \in [-\frac{1}{2}\pi,\frac{1}{2}\pi]$, and then flip the predicted ${p}_{gen}$ 
to obtain the final predicted body pose.

Then we can generate images from $z$ and $c_{gen}$ by:
\begin{equation}
	\begin{split}
    \label{eqn: generate_new}
     I_{gen} = G(z, c_{gen}, \Gamma_G (z, c_{gen})) \sim P(I_{gen} \vert z, c_{gen}, \Gamma_G (z, c_{gen})).
     \end{split}
\end{equation}
{In Sec. \ref{sec: EG3D Backbone}-\ref{sec: Deformation Module}}, 
we will introduce how our generator renders an image ${I}_{gen}$ with certain camera parameters ${c}_{gen}$ and a body pose ${p}_{gen}$ in detail.

  

\subsubsection{Backbone}
\label{sec: EG3D Backbone}
As shown in Fig. \ref{fig: pipeline},  we utilize EG3D \cite{DBLP:conf/cvpr/ChanLCNPMGGTKKW22} as our backbone. 
After predicting body pose ${p}_{gen}$ from latent code $z$ and camera parameters $c_{gen}$, we input $z$ into the mapping network, where ${p}_{gen}$ and $c_{gen}$ serve as its conditional inputs.
The resulting $w$ latent code is then used to modulate both a main generator and a background generator. 
{The background generator synthesizes the background image $I_{gen}^{bg}$. The main generator synthesizes feature maps, which are then reshaped to certain 3D representations.}
In this paper, we use the ``tri-grid'' 3D representation proposed by PanoHead, which helps alleviate mirror feature artifacts.
We assume that our tri-grids are ``normalized'' and canonical. 
This indicates that if we directly render the results using the tri-grids, we will obtain a canonical human body geometry representation with a neutral body pose.

{
During volume rendering, given a ray $\mathrm{r}(t) = \mathrm{o} + t\mathrm{d}$ pointing from its origin $\mathrm{o}$ (camera center) into direction $\mathrm{d}$, we sample point $\bold{x} = \mathrm{r}(t) = (x,y,z)$ on the ray and project $\bold{x}$ onto the tri-grid as $(x,y),(y,z),(z,y)$. 
These projections are used to sample features from the tri-grid, 
which are then fed into a decoder to obtain the color $f(\mathrm{r}(t))$ and density $\sigma(\mathrm{r}(t))$ of point $\bold{x}$ and to perform volume rendering: 
\begin{equation}
	\begin{split}
    \label{eqn: volume_rendering}
      I_{gen}'^{-}(\mathrm{r}) &= \int^{t_f}_{t_n} w(t) f(\mathrm{r}(t)) dt, \\
      I_{gen}^{seg} (\mathrm{r}) &= \int^{t_f}_{t_n} w(t)  dt, \\
        w(t) &=  \mathrm{exp} \left (- \int^{t}_{t_n} \sigma(\mathrm{r}(s))ds \right )\sigma (\mathrm{r}(t)),
     \end{split}
\end{equation}
where $t_n$ and $t_f$ {respectively denote} the near and far bounds along $\mathrm{r}$, $I_{gen}'^{-}$ is the rendered image, {and} $I_{gen}^{seg}$ is the rendered foreground mask.

Similar to PanoHead, the rendered image $I_{gen}'^{-}$ and the background image $I_{gen}^{bg}$ are composed using the foreground mask $I_{gen}^{seg}$, getting a composed raw image:
\begin{equation}
	\begin{split}
    \label{eqn: composed raw image}
      I_{gen}^{-} = (1-I_{gen}^{seg})I_{gen}^{bg} + I_{gen}'^{-}.
     \end{split}
\end{equation}
After that, 
$I_{gen}^{-}$ is fed to a super-resolution network to obtain the final high-resolution image $I_{gen}^{+}$.
}

\subsubsection{Deformation Module}
\label{sec: Deformation Module}
In Sec. \ref{sec: EG3D Backbone}, we assume that the tri-grids generated by our generator are canonical. 
Given a body pose $p_{gen}$, to achieve a final portrait that conforms to $p_{gen}$, we utilize the deformed mesh $M(p_{gen})$ to produce a deformation field that maps each sampled point $\bold{x} = (x, y, z)$ in the observation space to a corresponding point $\bold{x}' = (x', y', z') = (x + \Delta x, y + \Delta y, z + \Delta z) $ in the canonical space, where $ \Delta \bold{x} =  (\Delta x, \Delta y,  \Delta z) $ denotes the deformation field value.

{We draw inspiration from RigNeRF \cite{DBLP:conf/cvpr/AtharXSSS22}, which uses a mesh-guided deformation field and a residual deformation network to achieve full control of neural 3D portraits. To improve the training efficiency of our model, we exclude the NN-based residual deformation training and directly use the readily available mesh-guided deformation field. Specifically, we use a canonical SMPL mesh $M(0)$, where the pose is neutral ($0$ refers to neutral body pose), and the deformed SMPL mesh $M(p_{gen})$ to compute a deformation field.
}
Similar to RigNeRF, the SMPL deformation field value at a point $\bold{x}$ on the ray is defined as follows:
\begin{equation} 
	\begin{split}
    \label{eqn: original SMPLDef}
    & \Delta \bold{x}  = SMPLDef(\bold{x}, p_{gen}) = \frac{SMPLDef(\hat{\bold{x}},p_{gen})}{\exp(\Vert\bold{x} , \hat{\bold{x}}\Vert^2)}, \\
    & SMPLDef(\hat{\bold{x}},p_{gen}) = \hat{\bold{x}}_{M(0)} - \hat{\bold{x}}, \enspace
     \end{split}
\end{equation}
where $\hat{\bold{x}}$ is the closest point on the deformed mesh $M(p_{gen})$ to $\bold{x}$, and  $\Vert\bold{x} , \hat{\bold{x}}\Vert^2$ is the Euclidean Distance between $\bold{x}$ and $\hat{\bold{x}}$. $\hat{\bold{x}}_{M(0)}$ is the position of point $\hat{\bold{x}}$ on $M(0)$.

However, the deformation field in RigNeRF may encounter issues when the NN-based non-rigid deformation is disabled and the body pose $p_{gen}$ is large. 
The computation of RigNeRF's deformation field $\Delta \bold{x}$ depends only on the translation of the point, ignoring the relative positioning between the sample point and the mesh. This approach produces an ``offset face'' as depicted in Fig. \ref{fig: ablation_deform}.

To tackle this issue, as shown in Fig. \ref{fig: pipeline}, we utilize a deformation field that accounts for the positional relationship between $\bold{x}$ and its nearest face on the mesh:
\begin{equation}
\begin{split}
\label{eqn: new SMPLDef}
& \Delta \bold{x}  = SMPLDef(\bold{x},p_{gen}) = \left\{
\begin{aligned}
&\check{\bold{x}} - \bold{x}, \enspace \quad  \Vert\bold{x}, \hat{f}\Vert^2<\alpha\\
&0, \enspace \quad \Vert\bold{x}, \hat{f}\Vert^2 \geq \alpha \\
\end{aligned}
\right. \\
&\bold{x}  = C_{\hat{f}}(u,v,h) ,\quad  \check{\bold{x}}  = C_{\hat{f}}^{M(0)}(u,v,h), \enspace 
\end{split}
\end{equation}
where $\hat{f}$ is the face on $M(p_{gen})$ that is closest to $\bold{x}$, and  $ \Vert\bold{x}, \hat{f}\Vert^2$ denotes the Euclidean distance between $\bold{x}$ and  $\hat{f}$, $\alpha$ is a hyper-parameter that controls the ``thickness'' of the geometry beyond the mesh (we empirically set $\alpha$ as 0.25).

We first obtain the local coordinate system $C_{\hat{f}}$ of $\hat{f}$ using its vertices, which yields the local coordinates $(u,v,h)$ of $\bold{x}$ in $C_{\hat{f}}$.  Next, we obtain the local coordinate system $C_{\hat{f}}^{M(0)}$ of the same face on the template mesh $M(0)$ and use $(u,v,h)$ to compute the new global coordinates $\check{\bold{x}}$. 
In other words, if $\bold{x}$ is close to the mesh, its position relative to its closest face on the mesh remains unchanged.

\subsection{Losses}
\label{sec: Losses}

\subsubsection{Discriminator Loss}
We define the loss of the discriminator as:
\begin{equation}
\begin{split}
    \label{eqn: D discriminator loss}
    L_D = & -\mathbb{E} [\log(1-D(I_{gen}\vert c_{gen}, \hat{p}_{gen} ))] \\
          & - \mathbb{E} [\log (D(I_{real}\vert c_{real}, \hat{p}_{real} ) ] \\
          &  +  \lambda \mathbb{E} [ ||\nabla_{I_{real}} D(I_{real}\vert c_{real}, \hat{p}_{real} ) ||_2] + \lambda_p L_{p},
    \end{split}
\end{equation}
where $\lambda \mathbb{E} [ ||\nabla_{I_r} D(I_{real}\vert c_{real}, \hat{p}_{real} ) ||_2]$ is the gradient penalty, $\lambda_{p}$ 
represents the weight of body pose loss $L_{p}$. 
{We use $L_p$ to optimize the pose predictor $\Gamma_D$ in the discriminator as:}
\begin{equation}
	\begin{split}
    \label{eqn: body pose loss}
    L_{p} & = L_2({p}_{gen}, \hat{p}_{gen}),
     \end{split}
\end{equation}
where ${p}_{gen}$ could be regarded as the ground-truth body pose of $I_{gen}$ (since ${p}_{gen}$ is used to perform deformation), and $\hat{p}_{gen}$ is the body pose that is predicted by the discriminator from $I_{gen}$, $L_2$ denotes the $L_2$ distance.



\subsubsection{Generator Loss}
We define the generator's loss as follows:
\begin{equation}
\begin{split}
    \label{eqn: G discriminator loss}
    L_G = & - \mathbb{E} [\log (D(I_{gen}\vert c_{gen}, \hat{p}_{gen} )) ] \\
    & + \lambda_{preg} L_{preg},
    \end{split}
\end{equation}
where $L_{preg}$ represents the body pose regularization loss, and $\lambda_{preg}$ represents the weight of the regularization loss.
The body pose regularization loss $L_{preg}$ will be only employed in the very early stage of our training process see more details in Sec. \ref{sec: Training Details}).

 \subsubsection{Body Pose Regularization Loss}
 \label{sec: Body Pose Regularization Loss}
Although our network can learn the relative body pose between different images, it struggles to predict the absolute body pose due to the lack of prior information. The predicted pose can be viewed as a ``deviated'' pose, which has an offset from the true value. Since the camera system can be globally rotated, this offset does not affect camera parameters prediction (as in Pof3D \cite{shi2023pof3d}). However, we use a
SMPL model in the canonical space, which means a ``deviated'' body pose will result in an unnatural mesh deformation. To address this issue, we propose using $L_{preg}$ to constrain the value of the predicted pose as follows:
\begin{equation}
	\begin{split}
    \label{eqn: body pose reg loss}
    {p}_{gen} &= \Gamma_G (z, c_{gen}), \\
     L_{preg} & =  L_2({p}_{gen}, p_{coarse}), 
     \end{split}
\end{equation}
where $p_{coarse}$ represents the coarse body pose in the \textit{$\it{360}^{\circ}$PHQ} dataset, 
and $c_{gen}$ and $p_{coarse}$ are from the same real image.

\subsubsection{Overrall Loss}
Our full objective function is:
\begin{equation}
\begin{split}
    \label{eqn: full objective}
    L =  L_D + L_G 
    \end{split}.
\end{equation}

\begin{table}[t]
\caption{The training details of our three-stage training. M means million images. 
}

\scalebox{0.9}{
\begin{tabular}{@{}cccccc@{}}
\toprule
                         & Image                & $L_{preg}$        & Neural Rendering Resolution     & $\Gamma_G$              \\ \midrule
\multirow{2}{*}{Stage 1} & 0$\sim$0.2M          & \CheckmarkBold    & $64^2$                          & training          \\
                         & 0.2M$\sim$6M       & \XSolidBrush        & $64^2$                            & training        \\  \midrule
Stage 2                  & 6M$\sim$10M          & \XSolidBrush      & $64^2$                          & freeze           \\ \midrule
\multirow{2}{*}{Stage 3} & 10M$\sim$11M         & \XSolidBrush      & increase from $64^2$ to $128^2$ & freeze       \\ 
                         & 11M$\sim$13M         & \XSolidBrush      & $128^2$                         & freeze       \\\bottomrule
\end{tabular}
}
  \label{tab: training_details}
\end{table}

{
 \subsection{Training Details}
\label{sec: Training Details}
Our model is trained on 8 NVIDIA A40 GPUs. It takes 7 days to train our full model.
Taking into account the computational cost of the deformation field, we retain only the SMPL faces that fall within the bounding box of the volume rendering. 
The resolution of the training dataset is $256^2$. 
The body pose predictors $\Gamma_G$ and $\Gamma_D$ are composed of fully-connected layers with leaky ReLU as the activation functions. 

The proposed training strategy for our 3DPortraitGAN is composed of three stages, as shown in Tab. \ref{tab: training_details}.

\paragraph{Stage 1 - Warm Up} 
    The first stage is the warm-up period, from 0 to 6M images.
    We employ the regularization loss $L_{preg}$ during the initial phase of this stage. 
    Specifically, we employ 
    $L_{preg}$ 
    to the first 0.2M images while linearly decaying $\lambda_{preg}$ from 0.5 to 0 over the subsequent 0.2M images. This helps prevent the coarse body pose from negatively affecting the entire training process. 
    %
    %
    We utilize the swapping regularization method proposed by EG3D \cite{DBLP:conf/cvpr/ChanLCNPMGGTKKW22}. Specifically, we begin by randomly swapping the conditioning pose of $G$'s mapping network with another random pose with 100\% probability {and then linearly decay the swapping probability} 
    to 70\% over the first 1M images. For the remainder of this stage, we maintain a 70\% swapping probability. The neural rendering resolution remains fixed at $64^2$.

\paragraph{Stage 2 - Freeze $\Gamma_G$} 
    From 6M to 10M images, we encounter issues with the collapse of $\Gamma_G$ during training. 
    Therefore, we freeze $\Gamma_G$ while continuing training 3DPortraitGAN for 10M images.
    We randomly swap the conditioning pose of $G$'s mapping network with another random pose with 70\% probability during this stage, and {fix the} neural rendering resolution 
    at $64^2$.

\paragraph{Stage 3 - Increase Neural Rendering Resolution}  
    This stage represents the training from 10M images to 13M images of our full training process.
    In this stage, we gradually increase the neural rendering resolution of 3DPortraitGAN while {keeping the} other training settings 
    identical to those of Stage 2.
    Specifically, we linearly increase the neural rendering resolution from $64^2$ to $128^2$ from 10M to 11M images {and then} 
    retain the neural rendering resolution as $128^2$ until finishing training.
    We randomly swap the conditioning pose of $G$'s mapping network with another random pose with 70\% probability during this stage.

}

\begin{figure*}[t]
  \centering
  \includegraphics[width=\linewidth]{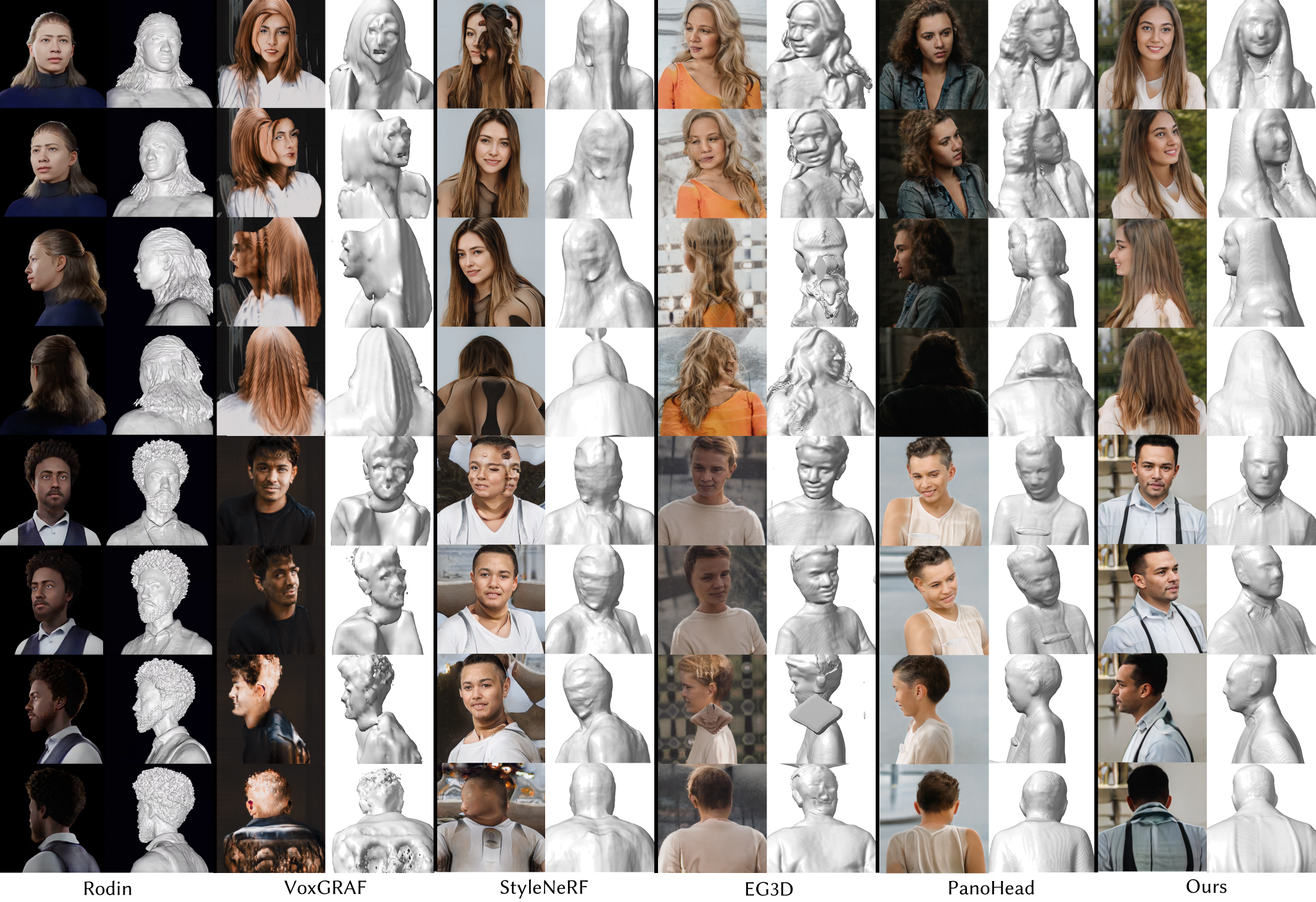}
  
  \caption{Qualitative comparison to state-of-the-art approaches. From left to right: Rodin \cite{DBLP:journals/corr/abs-2212-06135}, VoxGRAF \cite{DBLP:conf/nips/SchwarzSNL022}, StyleNeRF \cite{DBLP:conf/iclr/GuL0T22}, EG3D \cite{DBLP:conf/cvpr/ChanLCNPMGGTKKW22}, PanoHead \cite{An_2023_CVPR}, and our method. 
  To ensure a fair comparison, we randomly sample four camera parameters from our \textit{$\it{360}^{\circ}$PHQ} dataset and apply the same four camera parameters to all models in the generation while we set the body pose to the neutral body pose for our results. 
  {
  Since the results of Rodin are extracted from the official demo video, we do not apply the same camera parameters to them as we do to other models.
  }
  }
  \label{fig: comparison}
\end{figure*}

\section{Results}
\label{sec: Results}

Fig. \ref{fig: generation-2} displays a selection of random samples generated by our model. Real images from the \textit{$\it{360}^{\circ}$PHQ} dataset are randomly sampled (1st col) and passed through $\Gamma_D$ to predict the 
body poses. Latent codes are then randomly sampled, and the camera parameters and predicted body poses of the real images are used to generate images (2nd-7th cols).
We also show results rendered from steep camera parameters in Figs. \ref{fig:yaw}-\ref{fig:pitch}.

To illustrate 3DPortraitGAN's performance in generating novel views for real images, we perform latent code optimization in the $W$ latent space to real images using our 3DPortraitGAN model. The input real images are never seen during training. As shown in Fig.~\ref{fig: inversion}, 3DPortraitGAN produces reasonable reconstructed portrait geometry and appearance. See more discussion about our results in Sec. \ref{sec: Discussion}. {Further details on the inversion process can be found in the supplementary file.}

\begin{figure*}[t]
  \centering
  \includegraphics[width=\linewidth]{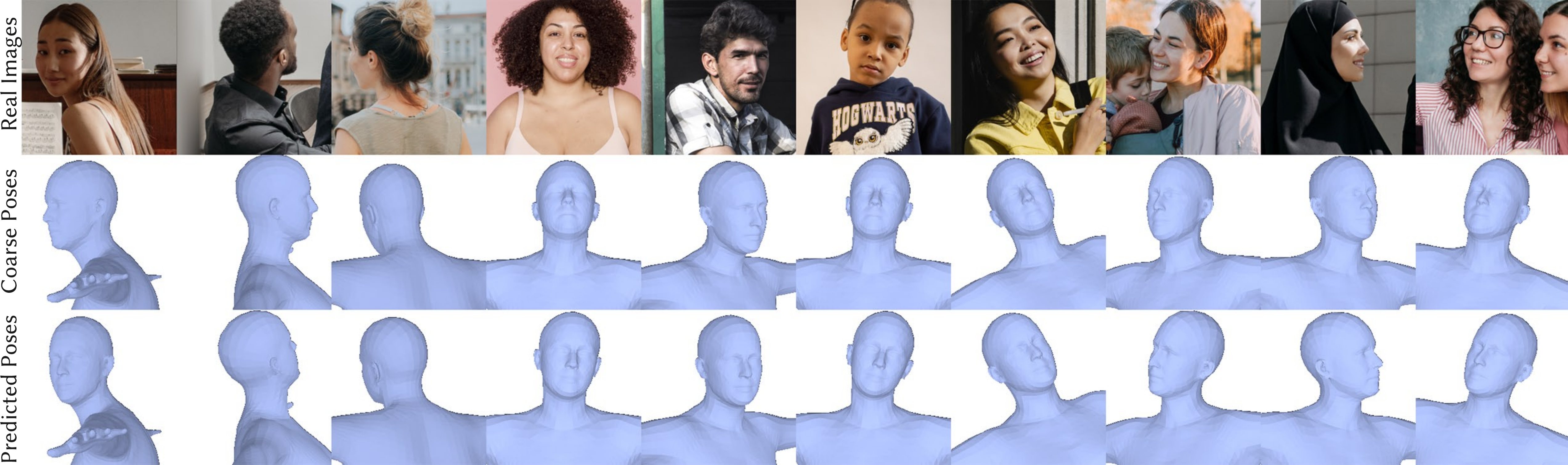}
  
  \caption{
  Comparison between the coarse body poses derived from our \textit{$\it{360}^{\circ}$PHQ} dataset (2nd row) and the body poses 
  predicted by our body pose-aware discriminator $\Gamma_D$  (3rd row) based on the real images (1st row).  }
  \label{fig: pose_predict}
\end{figure*}

\section{Comparison}
\label{sec: Comparison}
To be consistent with the scope of our method, we choose VoxGRAF \cite{DBLP:conf/nips/SchwarzSNL022}, StyleNeRF \cite{DBLP:conf/iclr/GuL0T22}, EG3D \cite{DBLP:conf/cvpr/ChanLCNPMGGTKKW22}, and PanoHead \cite{An_2023_CVPR} as state-of-the-art representations of the previous 3D-aware generators to compare. We {re-train those works on our dataset using their official implementation.} 
Rodin \cite{DBLP:journals/corr/abs-2212-06135} has a similar objective as ours. However, Rodin requires a multi-view image dataset of avatars to fit tri-planes, which cannot be applied to our single-view dataset. Furthermore, the dataset and code of Rodin are not publicly available. Therefore, we only show random Rodin results in our qualitative comparison.

\subsection{Qualitative Comparison}
In Fig. \ref{fig: comparison}, we present a qualitative comparison between the SOTA methods and 3DPortraitGAN. The results show that VoxGRAF, EG3D, and PanoHead suffer from distortion and artifacts due to their attempt to directly learn 3D portraits from a dataset with diverse body poses. The results of StyleNeRF exhibit interpolated faces, which can be more visible in our supplementary video. Rodin generates rendering-style results that lack realism.
In contrast, 3DPortraitGAN shows an enhancement in terms of the quality of results. Our approach generates high-quality multi-view rendering results and reasonable 3D geometry.

\begin{table}[t]
\caption{FID and face identity for our model and SOTA methods. 
$^\star$ means the body poses used  
to generate images are predicted by $\Gamma_D$ from real images in \textit{$\it{360}^{\circ}$PHQ},  $^\diamond$ means the body poses used to generate images are sampled from the pose distribution predicted by $\Gamma_G$. 
}
\scalebox{0.85}{
\begin{tabular}{@{}ccccccc@{}}
\toprule
             & VoxGRAF &StyleNeRF  & EG3D  &PanoHead & Ours                             \\ \midrule
FID $\downarrow$          & 39.41   &22.16      & 14.52 & 14.49   &\textbf{12.88}$^\star$ /\textbf{13.75}$^\diamond$    \\ \midrule 
Identity $\uparrow$    &  0.52   &0.46       & 0.57  & 0.60  & \textbf{0.68}  \\ 
\bottomrule
\end{tabular}
}
  \label{tab: metrics}
\end{table}

\subsection{Quantitative Comparison}
Regarding the quantitative results, we compare our method to SOTA alternatives using Fr\'echet Inception Distance (FID) \cite{DBLP:conf/nips/HeuselRUNH17} and facial identity metrics (refer to Tab. \ref{tab: metrics}).

To assess the rendering quality of models, we use FID, which computes the distance between the distribution of the generated images and that of the real images to evaluate the quality and diversity of the generated images.
For each model, we generate 50K images using the camera parameters sampled from the \textit{$\it{360}^{\circ}$PHQ} dataset. 
For our 3DPortraitGAN, we utilize pose predictor $\Gamma_D$ to predict body poses from real images in \textit{$\it{360}^{\circ}$PHQ} ($^\star$ in Tab. \ref{tab: metrics}), and also sample body poses from the pose distribution 
predicted by the pose predictor $\Gamma_G$ in the generator ($^\diamond$ in Tab. \ref{tab: metrics}). 
%
Our 3DPortraitGAN model shows notable improvements in FID.
Moreover, we obtain similar FID scores when utilizing the body poses predicted by both $\Gamma_D$ and $\Gamma_G$ in our model, implying that the  body pose distribution predicted by $\Gamma_G$ closely resembles the genuine distribution.
We employ ArcFace \cite{DBLP:conf/cvpr/DengGXZ19} to evaluate the ability of models to maintain facial identity. Specifically, we produce a pair of images with different camera views for 1,024 randomly selected latent codes for each model.
We observe that the performance of ArcFace is affected by extreme camera angles, in which the face is completely occluded. To address this, we solely sample camera positions for each view with $\mu \in[0.25\pi,0.75\pi]$ from the \textit{$\it{360}^{\circ}$PHQ} dataset, ensuring unobstructed face regions.
To ensure a fair comparison, we set the body pose to a neutral position for our model and set the conditional camera parameters of all models as the average camera parameters. We also set the truncation of all models as 0.6.
As shown in Tab. \ref{tab: metrics}, our model presents improvements in facial identity consistency, indicating that our model attains superior performance in generating realistic view-consistent results.

\begin{table}[t]
\caption{ 
 We compare the performance of our 3DPortraitGAN model against the baseline model by generating 1,024 portrait images without performing deformation and computing the mean and standard deviation values for the predicted body poses.
}
\begin{tabular}{@{}ccc@{}}
\toprule
                    & w/o pose learning  & Ours     \\ \midrule
mean                &  3.19$^\circ$             & \textbf{2.51}$^\circ$    \\  
standard deviation  &  2.88$^\circ$             & \textbf{2.49}$^\circ$    \\ \bottomrule
\end{tabular}
  \label{tab: ablation Pose prediction}
\end{table}

\section{Ablation Study}
\label{sec: Ablation Study}

\begin{figure*}[t]
  \centering
  \includegraphics[width=\linewidth]{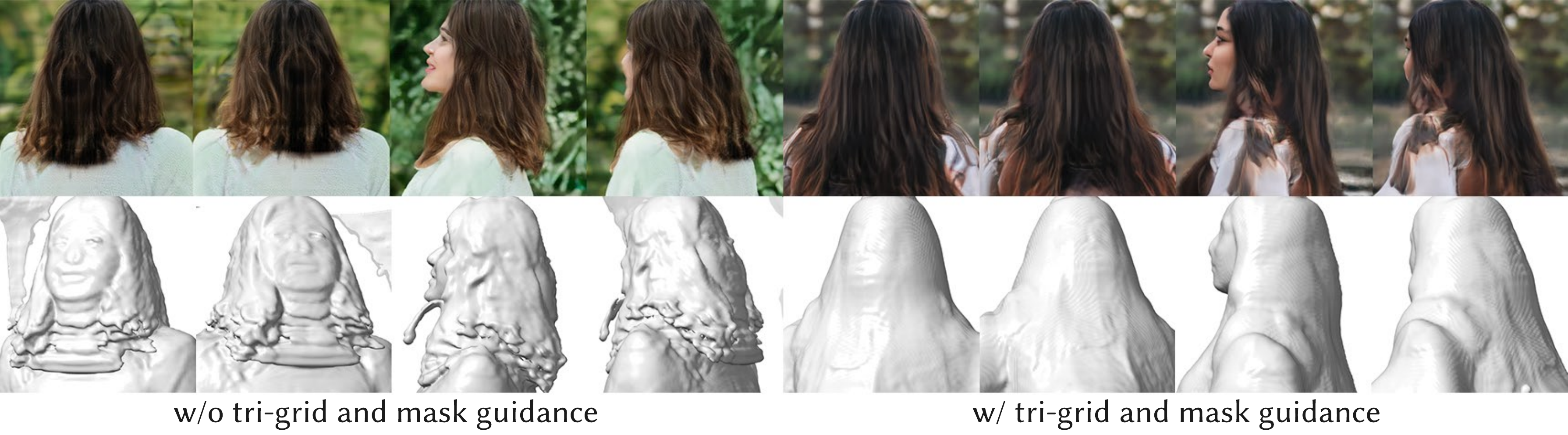}
  
  \caption{The tri-grid and mask guidance is crucial for eliminating the face artifact on the back of the head.}
  \label{fig: rear-view-reg}
\end{figure*}

\subsection{Pose prediction}
\label{sec: ablation pose prediction}
As illustrated in Fig. \ref{fig: pose_predict}, we present the coarse body poses from \textit{$\it{360}^{\circ}$PHQ} alongside the body poses predicted by our body pose-aware discriminator. Our results exhibit a notable enhancement in the accuracy of the predicted body poses, surpassing the precision of the coarse body poses acquired from real images through an off-the-shelf body reconstruction approach \cite{DBLP:conf/cvpr/ChoiMPL22}.

To demonstrate that our pose predictor generates more accurate body poses and improves model performance, we conduct an experiment by directly utilizing the coarse body poses from \textit{$\it{360}^{\circ}$PHQ} to train a baseline model. In particular, we remove the pose predictors from the generator and discriminator and train the model using the coarse body poses obtained from \textit{$\it{360}^{\circ}$PHQ}.
For both baseline and 3DPortraitGAN models,
we generate 1,024 portrait images by using randomly selected latent codes, camera parameters with $\mu = \frac{1}{2} \pi$ and $\nu = \frac{1}{2} \pi$ (i.e., frontal view), and neutral body pose (i.e., no deformation performed). 
To compare the performance of the two models in generating the canonical 3D representation, we utilize $\Gamma_D$ from our full model to predict the body poses of the randomly generated portraits. Then we calculate the mean and standard deviation of the predicted body poses' absolute values for each model.

As outlined in Tab. \ref{tab: ablation Pose prediction}, our 3DPortraitGAN model produces lower mean and standard deviation values for body poses when the deformation is disabled. This indicates that our pose learning and prediction module aids the generator in better learning the canonical 3D representation.
The higher body pose mean and deviation observed in the baseline model may be due to some distortion cases in the random outputs. 
As coarse body poses are used as conditional labels for the discriminator to compute image scores and for the generator to compute the deformation field, their inaccuracy will lead the generator to learn some distorted samples to fit the real image distribution.
We 
present some distorted results in the supplementary file.

{\subsection{Tri-grid and Mask Guidance}
We observe the presence of a face in the rear view (refer to Fig. \ref{fig: rear-view-reg}), which we call a ``rear-view-face'' artifact. Despite the generator's ability to learn the texture of hair at the back of the head, the face geometry is sometimes apparent in the shape extracted by the marching cubes algorithm. Such an artifact also exists in the comparison results in Rodin \cite{DBLP:journals/corr/abs-2212-06135} and PanoHead \cite{An_2023_CVPR}, where they retrain EG3D using their own dataset.
We attribute this phenomenon to the geometric ambiguity when using single-view images as training data. The discriminator can only assess the rendered images, leading to an under-determined problem in characterizing the geometry of the back of the head. In the absence of constraints, our model may converge to a suboptimal solution marked by front-back symmetric geometry.

To address this problem, in Sec. \ref{sec: Methodology}, we employ the tri-grid 3D representation and mask guidance proposed by PanoHead \cite{An_2023_CVPR}.
The tri-grid helps alleviate mirror feature artifacts, and the foreground masks provide more accurate 3D prior to some extent.
Fig. \ref{fig: rear-view-reg} demonstrates the effectiveness of tri-grid and mask guidance. When tri-grid and mask guidance are not applied (left), the generator simply learns a face geometry for the back of the head. Conversely, with 
tri-grid and mask guidance 
(right), the generator learns a smoother, more natural-looking geometry. 

}

\subsection{Mesh-guided Deformation Field}
In Sec. \ref{sec: Deformation Module}, we mentioned that the mesh-guided deformation field in RigNeRf \cite{DBLP:conf/cvpr/AtharXSSS22} would 
cause an ``offset face'' artifact. We show this phenomenon in Fig. \ref{fig: ablation_deform}. We deform the portraits using body pose $p_{n} = [0,\frac{\pi}{2},0]$ and $p_{h} = [0,0,0]$, it can be seen that the deformed portrait using RigNeRf deformation field suffer from the ``offset face'' artifact, while ours is more realistic.

\begin{figure}[t]
  \centering
  \includegraphics[width=\linewidth]{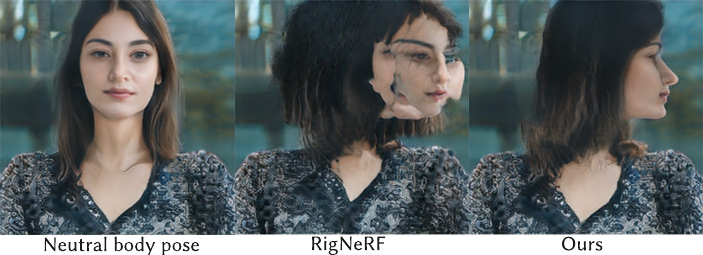}
  \caption{
  Comparison of mesh-guided deformation field in RigNeRf and ours. 
  }
  \label{fig: ablation_deform}
\end{figure}

                                                                                                                                                                                                                                                                                                                                                                                                                           
\section{Discussion}
\label{sec: Discussion}

Despite its noteworthy performance, our method has some limitations. 
Firstly, we employ a deformation field solely governed by the SMPL model, which does not account for the shape of the resulting geometry of a portrait. Although this deformation method does not interfere with our training process, it leads to some artifacts during inference, such as truncated long hair {(highlighted by the green box in Fig. \ref{fig: limitation} (a))}. Additionally, the computation of the deformation field is time- and memory-intensive, making our training much slower than EG3D and costing almost twice as much time.

Secondly, the pose predictor $\Gamma_G$ within the generator is prone to collapsing and heavily influences the entire framework during the training phase.  To prevent this, we freeze $\Gamma_G$ after it is trained for a certain duration. 
Furthermore, while the pose self-learning strategy in our framework can assist in predicting more precise body poses and alleviate distortions in the results, as shown in Tab. \ref{tab: ablation Pose prediction}, our model still cannot achieve entirely canonical representations {(see distorted samples in Fig. \ref{fig: limitation} (c))}. This is because of the inaccurate predictions in our pose learning method, even if the outcome is better than the coarse body poses in our dataset.
During training, we directly used the camera parameters in our \textit{$\it{360}^{\circ}$PHQ} dataset as training labels. However, the inaccurate estimation of these camera parameters might cause geometric artifacts. This issue can be overcome by using a more accurate body pose estimation method in the future.

{
Thirdly, we employ tri-grid and foreground mask guidance to alleviate the ``rear-view-face'' artifacts. However, as the rear-view data in our dataset is significantly less than the frontal data, our approach does not entirely eradicate them (highlighted by the blue boxes in Fig. \ref{fig: limitation}(b)). This issue might be addressed by adding more rear-view data to the training set.
}

Finally, our model still lacks the expressive power to accurately represent real-life portrait images, as seen in the real image inversion results presented in Fig. \ref{fig: inversion}. 
{We attribute this to the limited resolution ($256^2$) of the tri-grids and the training images. This issue could be addressed by increasing the resolution and seeking more efficient training strategies to prevent the model from consuming excessive computation to train high-resolution tri-grids and final results.}

\begin{figure}[t]
  \centering
  \includegraphics[width=\linewidth]{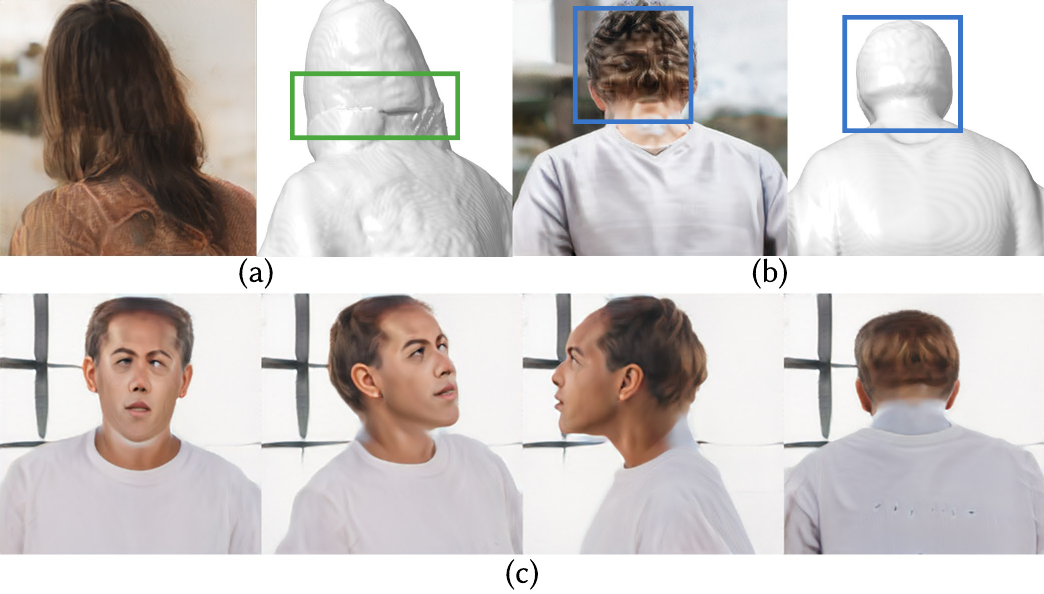}
  \caption{
  Limitations of our method. (a) Our deformation field causes artifacts in hair rendering (highlighted by the green box). (b) The ``rear-view-face'' artifacts on the occiput are still present (highlighted by the blue boxes). {(c) Distorted results of our method.}
  }
  \label{fig: limitation}
\end{figure}
\section{Conclusion}
\label{sec: Conclusion}

This paper presented 3DPortraitGAN, the first 3D-aware one-quarter headshot portrait generator that can learn canonical 3D avatar distribution images from a collection of single-view real portraits with diverse body poses. 
We started by developing the first large-scale dataset of high-quality, single-view real portrait images with 
a wide range of camera parameters and body poses.
Then, based on the desired camera views, we used a mesh-guided deformation module to generate portrait images with specific body poses. This enables our generator to learn a canonical 3D portrait distribution while accounting for body pose.
In addition, we incorporate two pose predictors into our framework: one controls the generator's learning of the body pose distribution and the other assists the discriminator in predicting accurate body poses from input images. We see our work as an exciting step forward in multi-view portrait generation, and we hope it will inspire future research{, such as talking heads, realistic digital humans, and 3D avatar reconstruction.} 

\begin{figure*}[t]
  \centering
  \includegraphics[height=0.96\textheight]{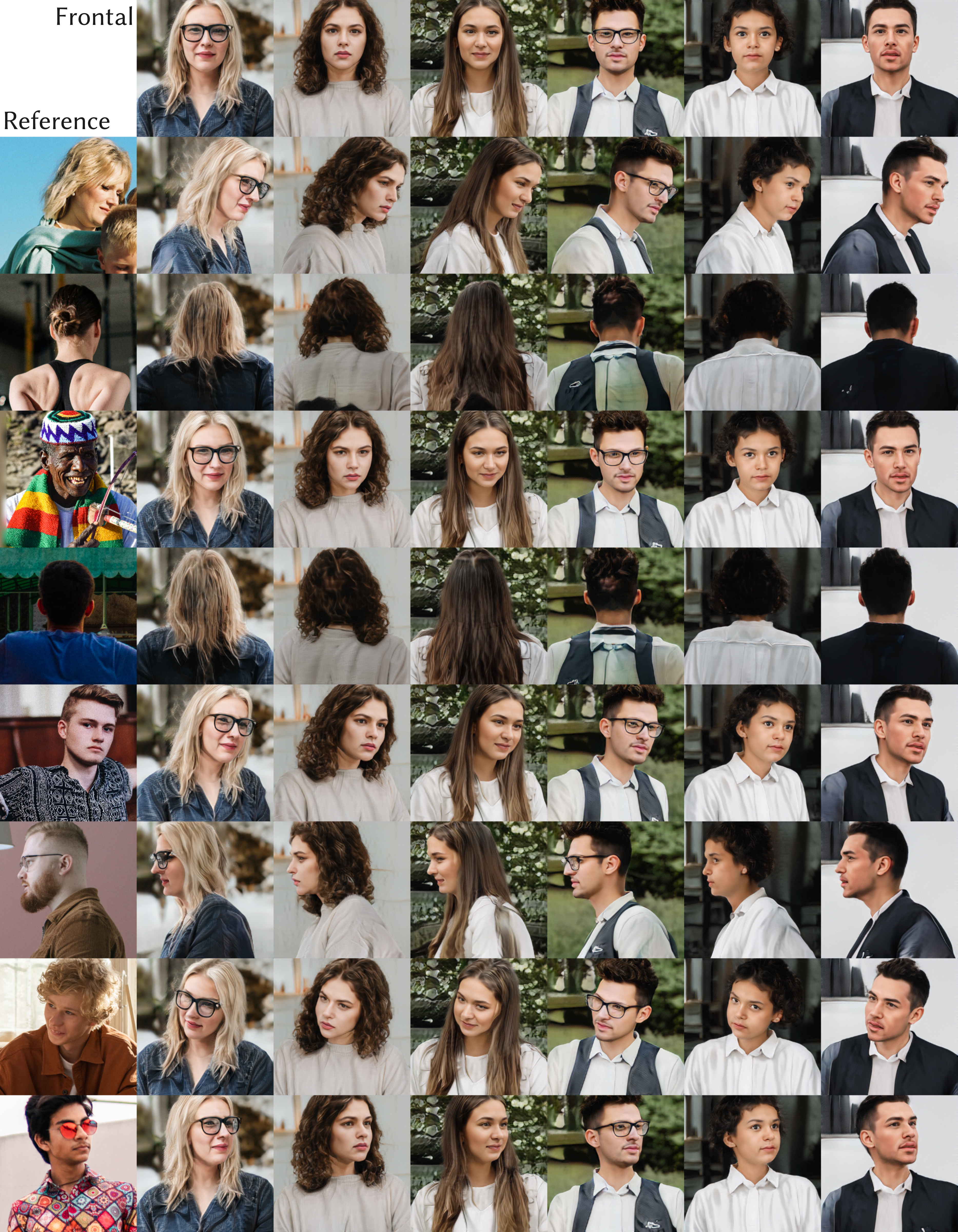}
  
  \caption{
  Randomly posed examples at $256^2$. We apply truncation  with $\psi=0.6$. The generator is conditioned on the average camera parameters and neutral body pose. The synthesized images are rendered from the camera parameters of the reference images while the body poses are predicted from the reference images by using
  our body pose-aware discriminator.
  }
  \label{fig: generation-2}
\end{figure*}

 \begin{figure*}[h]
            \centering
            \includegraphics[width=0.95\textwidth]{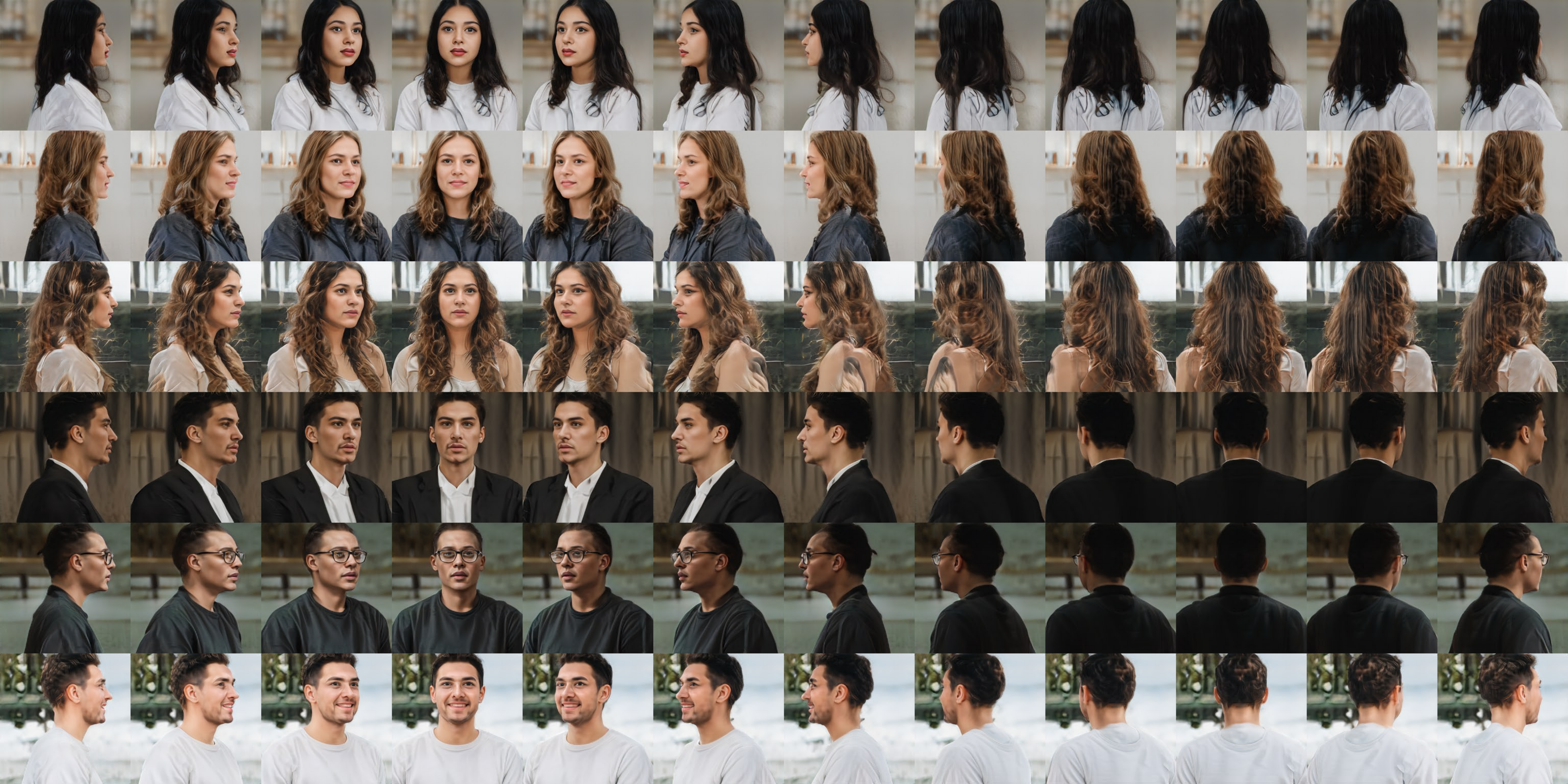}
                  \caption{Extrapolation to steep yaw angles, with truncation $\psi=0.5$.}
                  \label{fig:yaw}
        \end{figure*}

 \begin{figure*}[h]
            \centering
            \includegraphics[width=0.75\textwidth]{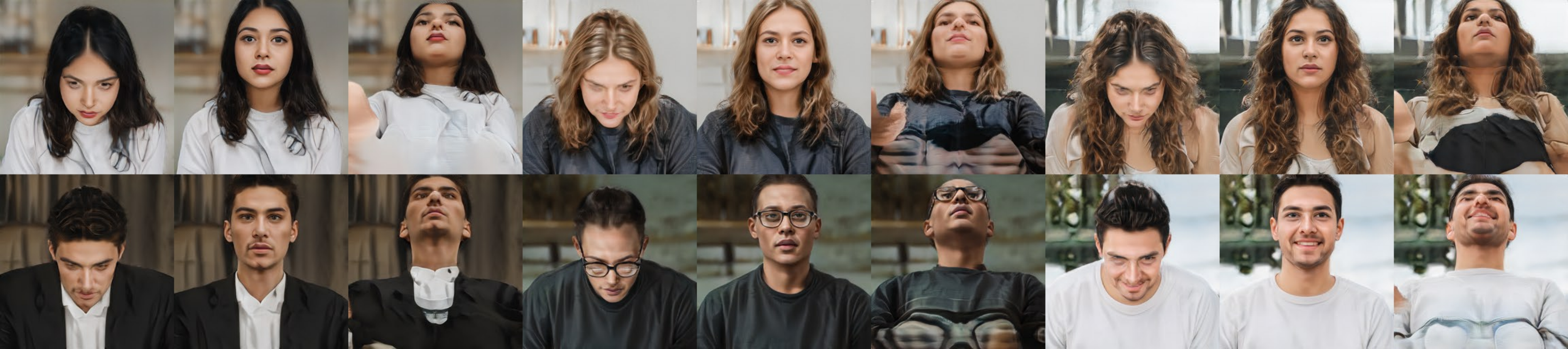}
                  \caption{Extrapolation to steep pitch angles, with truncation $\psi=0.5$.}
                  \label{fig:pitch}
        \end{figure*}

\begin{figure*}[t]
  \centering
  \includegraphics[width=0.96\textwidth]{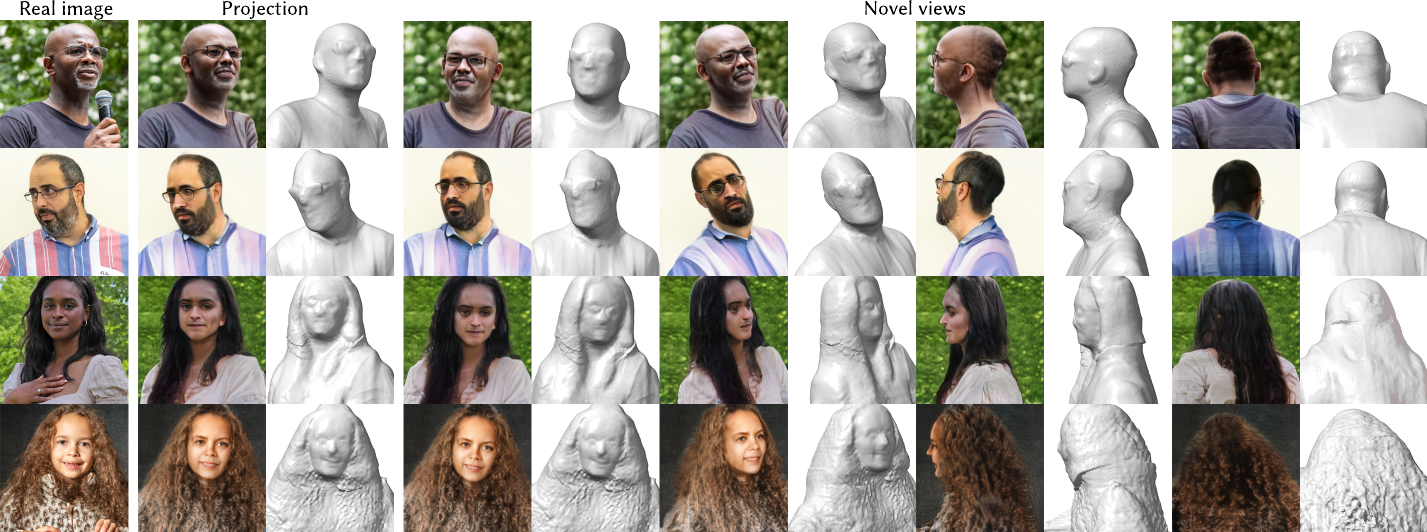}
  
  \caption{
  {To fit the single-view real image, we employ latent code optimization. 
  Specifically, for each real image (1st col), we perform latent code optimization to obtain the corresponding latent code in the $W$ latent space (2nd col). Then we use our model to render the obtained latent code using four novel views from the \textit{$\it{360}^{\circ}$PHQ} dataset. We utilize the body poses predicted by $\Gamma_D$ from the real images.
  The input real images are unseen during training.}
  }
  \label{fig: inversion}
\end{figure*}

\bibliographystyle{ACM-Reference-Format}
\bibliography{main-bibliography}

\end{document}